\documentclass[12pt]{article}
\usepackage{amsmath,amssymb,amsfonts}
\usepackage{graphicx}
\usepackage{xcolor}
\usepackage{hyperref}
\usepackage{caption}
\usepackage{subcaption}
\usepackage{float}
\usepackage{mathrsfs}
\usepackage{cite}
\usepackage{url}
\usepackage{microtype}
\usepackage{booktabs}
\usepackage{authblk}
\usepackage{enumitem}

\usepackage[margin=1in]{geometry}

\title{Geometric Analysis of Reasoning Trajectories: A Phase Space Approach to Understanding Valid and Invalid Multi-Hop Reasoning in LLMs}

\author{Javier Marín}
\affil{javier@jmarin.info}

\date{October 5, 2024}

\begin{document}

\maketitle

\begin{abstract}
This paper proposes a novel approach to analyzing multi-hop reasoning in language models through Hamiltonian mechanics. We map reasoning chains in embedding spaces to Hamiltonian systems, defining a function that balances reasoning progression (kinetic energy) against question relevance (potential energy). Analyzing reasoning chains from a question-answering dataset reveals that valid reasoning shows lower Hamiltonian energy values, representing an optimal trade-off between information gathering and targeted answering. While our framework offers complex visualization and quantification methods, the claimed ability to "steer" or "improve" reasoning algorithms requires more rigorous empirical validation, as the connection between physical systems and reasoning remains largely metaphorical. Nevertheless, our analysis reveals consistent geometric patterns distinguishing valid reasoning, suggesting this physics-inspired approach offers promising diagnostic tools and new perspectives on reasoning processes in large language models.

\vspace{0.5em}
\noindent\textbf{Keywords}: Multi-hop Reasoning, Reasoning Trajectories, Emergent Abilities, Mechanistic Interpretability, Hamiltonian Mechanics, Geometric Reasoning Analysis
\end{abstract}

\section{Introduction}
\subsection{Motivation for a physics-inspired approach}

The scientific method integrates mathematical abstractions with empirical observations to advance our understanding of natural laws \cite{popper1959}. Mathematical equations effectively encode physical processes \cite{wigner1990}, creating abstract structures that humans can manipulate intellectually \cite{penrose2006}. This approach generates predictions from theoretical models that can be experimentally verified.

The philosophical underpinnings include questions about abstract mathematical objects' existence beyond physical reality \cite{quine1948}, how we acquire knowledge of abstract domains \cite{goldman1967}, and structural realism's claim that mathematical and physical structures correspond \cite{worrall1989}. Universal structural realism proposes that the physical universe maps directly to mathematical structures \cite{tegmark2008}.

Physics' success in mathematically modeling complex phenomena suggests similar approaches could benefit artificial intelligence systems and cognitive science \cite{carleo2019}. Mathematical models serve as powerful tools for describing and predicting physical systems \cite{feynman1967}, capturing essential properties in forms suitable for analysis and application \cite{andersen1972}. The successes of physics-based formalisms in diverse fields \cite{barabasi1999, bialek2012} provide a solid justification for exploring their relevance in AI systems, particularly in the assessment and improvement of large language models' reasoning capabilities \cite{carleo2019, mehta2019}. Similar to how physicists apply mathematical models to explain the behavior of elementary particles, quantum fields, and particle states \cite{weinberg1995}, we propose the use of analogous formalisms for understanding the dynamics of LLMs reasoning within embedding spaces \cite{bengio2013, mikolov2013a}. This physics-based methodology provides a robust framework for analyzing LLMs complex reasoning, potentially yielding important advancements in the discipline \cite{wu2020}.

\subsection{Background on multi-hop reasoning in AI}

Multi-hop question-answering, where multiple facts are needed to derive an answer, is an important step to perform complex reasoning and provide explanations for answers in Language Models, LLMs \cite{yang2018}. QA provides a quantifiable and objective way to test the reasoning ability of intelligent systems. QA tasks provide numerical metrics such as accuracy, F1 score, or mean reciprocal rank, allowing for precise comparison between different AI systems. QA assignments generally have clearly defined correct answers, hence reducing subjectivity in evaluation and minimizing human bias in assessment. QA processes can be designed to evaluate several types of reasoning like deductive or inductive reasoning, deriving conclusions from established premises, and abductive reasoning (formulating the most plausible answer from partial knowledge) \cite{garbuio2021}.

Recent developments using knowledge graphs struggle to model multi-hop relations efficiently and lack transparency into the model's prediction rationale \cite{feng2020}. For example, knowledge graphs (KGs) neglect the latent relational information among question concepts and answers. \cite{dong2023} suggested a hierarchy-aware methodology that uses hierarchical structures in knowledge graphs to enhance comprehension and reasoning about complex questions, addressing the flaw of prior approaches that mostly emphasized language models for context encoding. This method captures more complex interconnections between concepts and facilitates a thorough understanding of semantic connections that may remain hidden in simpler representations. Hierarchical architectures facilitate advanced reasoning patterns that mimic human cognitive processes. It can more successfully address challenging questions requiring multi-step or multi-level reasoning \cite{wang2022}.

But we still have several challenges to improve reasoning processes. One of the most important requirements is LLM interpretability and explainability \cite{singh2024}. The struggle in understanding the internal reasoning process in LLMs, especially in deep neural networks, limits their real assessment and the capability of generating human-understandable explanations for AI decisions and inferences \cite{lipton2018, ribeiro2016}. Ensuring that reasoning processes are robust to slight variations in input or context, and developing models that can generalize reasoning skills across different domains and types of questions is another key target we must address to improve the reasoning process \cite{bengio2021, geirhos2020}. Without these efforts, we can't identify and correct biases in model's reasoning to ensure fair reasoning across different demographic groups or topic areas \cite{mehrabi2021}. We have also some challenges in scaling up reasoning capabilities to handle increasingly complex and multi-step problems and integrating updated external knowledge \cite{bender2021}. This is important to enhance thinking processes where model has to take decisions with limited information. \cite{jhamtani2020a, singh2024}. Addressing ambiguity in natural language questions and contexts will be decisive for improving the model's reasoning abilities, as will the development of new metrics that overcome existing ones' limits in capturing the nuances of complex reasoning. We must also deal with the reality of temporal and causal thinking, particularly in the modeling and analysis of temporal sequences and causal relationships \cite{maruthi2022}. Additional issues like ethical reasoning, adversarial attacks \cite{zhang2020}, long-term consistency, and human-AI collaboration \cite{jarvela2023} will require substantial revision to develop effective reasoning models.

An further important roadblock in enhancing the model's reasoning process is training data. Recent multi-hop question answering datasets seek to address several shortcomings of earlier multi-hop QA datasets \cite{ho2020, jhamtani2020b, welbl2018, yang2018}. These new datasets provide thorough elucidations of the reasoning process from QA, improving deficiencies in existing datasets. They additionally provide "evidence information" which defines a reasoning pathway for multi-hop questions, adding comprehensive explanations for predictions and simplifying model's reasoning abilities evaluation \cite{dua2019, khot2020}. Moreover, these datasets address the issue noted in earlier versions, where several examples lacked the necessity for multi-hop reasoning, by using logical rules to generate questions that are both natural and require multi-hop reasoning \cite{chen2019}. The mechanisms for generating question-answer pairs aim to guarantee both, the multi-hop reasoning and the quality of the questions. These datasets may improve the development of more transparent models in question answering and create a more rigorous standard for evaluating and developing multi-hop QA systems.

\section{Hamiltonian of dynamical systems}
\subsection{Brief review of Hamiltonian mechanics}

The Hamiltonian formalism is an elaborated mathematical framework for the analysis of conservative dynamical systems in classical mechanics. It should be noted that the majority of macroscopic physical phenomena observed in "classical" reality lack a Lagrangian description, thereby precluding a Hamiltonian formulation. Such physical systems are predominantly dissipative in nature, characterized by non-conservation of energy due to irreversible processes.
The presence of symmetries, which manifest as conserved quantities according to Noether's theorem \cite{kosmann2011}, can be rigorously derived from the Lagrangian formalism (or its Hamiltonian counterpart), with the existence of such a formalism being a necessary precondition. Fundamental physical theories have shown that nature's elementary interactions—gravitational, electromagnetic, strong nuclear, and weak nuclear forces—adhere to conservation principles and consequently take Lagrangian descriptions, allowing their formulation via the principle of stationary action. This formalism may be conceptualized as a geometric language permeating multiple domains of theoretical physics \cite{glattfelder2019}, aligned with Hilbert's proposition regarding the fundamentally geometric structure of physical theory \cite{young1998}.

The application of Hamiltonian mechanics to language model reasoning in embedding spaces offers two significant advantages. First, it provides a mathematically rigorous framework for characterizing the evolution of semantic representations across high-dimensional manifolds, potentially revealing conserved quantities in reasoning processes that may correspond to fundamental cognitive invariants (symmetries). Second, the symplectic structure  inherent in Hamiltonian systems \cite{goldman1984} enables the identification of canonical transformations that preserve the fundamental dynamics of reasoning while simplifying its analysis, potentially offering insights into the optimization of inferential pathways that might otherwise remain unseen in standard approaches to natural language processing.

We can define Hamiltonian with the following $2n$ ordinary differential equations \cite{easton1993}:

\begin{equation}
\dot{q} = H_p \, , \quad \dot{p} = -H_q
\end{equation}

\begin{equation}
\dot{q_i} = \frac{\partial H}{\partial p_i}(t, q, p) \, , \quad \dot{p_i} = -\frac{\partial H}{\partial q_i}(t, p, q)
\end{equation}
where $H = H(t, q, p)$ is the Hamiltonian, $q$ and $p$ are the position and momentum vectors of a mechanical system with $n$ degrees of freedom, and $t$ is the time. In these equations $(t, q, p) \in O$ -- an open set in $\mathbb{R}^1 \times \mathbb{R}^n \times \mathbb{R}^n$, $and$ $1 < i < n$.
Given the Lagrangian \cite{de2011}
\begin{equation}
L = T - U
\end{equation}
where $T$ is the kinetic energy and $U$ is the potential energy. We can rewrite this equation as 
\begin{equation}
L = T(q, \dot{q}) - U(q)
\end{equation}

\begin{equation}
\frac{d}{dt}\left(\frac{\partial L}{\partial \dot{q}}\right) = \frac{\partial L}{\partial q}
\end{equation}
Equation 5 is the Euler-Lagrange equation describing the motions of the system, and is equivalent to the Hamiltonian system \cite{hairer2006}:

\begin{equation}
H(q, p, t) = p^T \dot{q} - L(q, \dot{q}, t)
\end{equation}
Hamiltonian equations can be rewritten as
\begin{equation}
\dot{q} = \nabla_p H(q, p), \quad \dot{p} = -\nabla_q H(q, p)
\end{equation}

The Hamiltonian formalism introduces the concept of phase space, a $2n$-dimensional space where $n$ is the number of degrees of freedom. Each point in phase space represents a unique state of the system, defined by its position and momentum coordinates $(q, p)$. The phase space of a Hamiltonian system is a symplectic manifold, and Hamiltonian flows preserve the symplectic structure \cite{goldman1984}. Symplectic structures are fundamental geometric objects in differential geometry and classical mechanics \cite{goldman1984}. These objects are a framework for understanding the relationship between position and momentum in physical systems using Hamiltonian equations of motion. We could say that symplectic structures are specific rules that define how things move in physics, similar to an equation for motion. They help us grasp how items' positions and velocities are related, allowing us to predict how things will evolve over time.

Formally, a symplectic structure on a smooth manifold $M$ is a closed, non-degenerate 2-form $\omega$. This structure on an even-dimensional manifold $M$ satisfies two key properties \cite{goldman1984}:
\begin{enumerate}[label=\alph*)]
\item Closure: $d\omega = 0$, where $d$ is the exterior derivative.
\item Non-degeneracy: For each point $p$ in $M$, the map $v \mapsto \omega_p(v,\cdot)$ from the tangent space $T_p M$ to its dual is an isomorphism \cite{goldman1984}.
\end{enumerate}

These structures allow for the definition of Poisson brackets and canonical transformations \cite{easton1993}. In local coordinates $(q_i, p_i)$, a standard symplectic form can be written as $\omega = \sum_i dq_i \wedge dp_i$. In accordance with Liouville's theorem \cite{prugovecki1979}, these structures preserve phase space volume. while ensuring the conservation of certain geometric properties under the flow of Hamiltonian vector fields.

Symplectic structures and Poisson brackets (explained in section 2.2) are fundamental tools from classical mechanics that we can adapt to analyze reasoning paths in LLMs embedding spaces. By mapping reasoning chains to trajectories in a symplectic space, we can use these frameworks to quantify the ``energy'' and ``dynamics'' of cognitive processes, potentially uncovering underlying principles of effective reasoning and guiding the development of more robust algorithms.

\subsection{The Poisson bracket}

The fundamental feature of Hamiltonian systems is the conservation of energy. In isolated systems, the Hamiltonian system $H(q, p, t)$ remains constant over time, representing the ``total energy'' of the system:

\begin{equation}
\frac{dH}{dt} = \frac{\partial H}{\partial t} + \{H, H\} = \frac{\partial H}{\partial t} = 0
\end{equation}
where $\{,\}$ denotes the Poisson bracket operator \cite{karasev2012}. Many of the special properties of Hamiltonian systems are formulated in terms of the Poisson bracket operator. 

Let $F$, $G$, and $H$ be smooth functions from an open set $O$ in $\mathbb{R}^1 \times \mathbb{R}^n \times \mathbb{R}^n$, the Poisson bracket of $F$ and $G$ is defined as \cite{easton1993}:

\begin{equation}
\{F, G\} = \nabla F^T J \nabla G = \frac{\partial F^T}{\partial q} \frac{\partial G}{\partial p} - \frac{\partial F^T}{\partial p} \frac{\partial G}{\partial q}
\end{equation}
where $\{F, G\}$ is a smooth map from $O$ to $\mathbb{R}^1$. We can verify that $\{\cdot,\cdot\}$ is skew-symmetric and bilinear. When $H$ is independent of $t$, a critical point of $H$ as a function represents an equilibrium point of the Hamiltonian system's differential equations.

\subsection{Canonical transformations}

Canonical transformations are a central component in Hamiltonian mechanics. Generalized canonical transformations for generalized Hamiltonian systems transform one Hamiltonian system into another while maintaining its original structure \cite{fujimoto2001}. 

A set of transformations of $x$, $y$ and $H$, we say that is a canonical transformation for generalized Hamiltonian systems if it transforms the time-varying generalized Hamiltonian system into another one \cite{fujimoto2001}.

\begin{equation}
\bar{x} = \Phi(x,t)
\end{equation}
\begin{equation}
\bar{H} = H(x,t) + U(x,t)
\end{equation}
\begin{equation}
\bar{y} = y + a(x,t)
\end{equation}

\begin{figure}[h]
\centering
\includegraphics[width=0.75\textwidth]{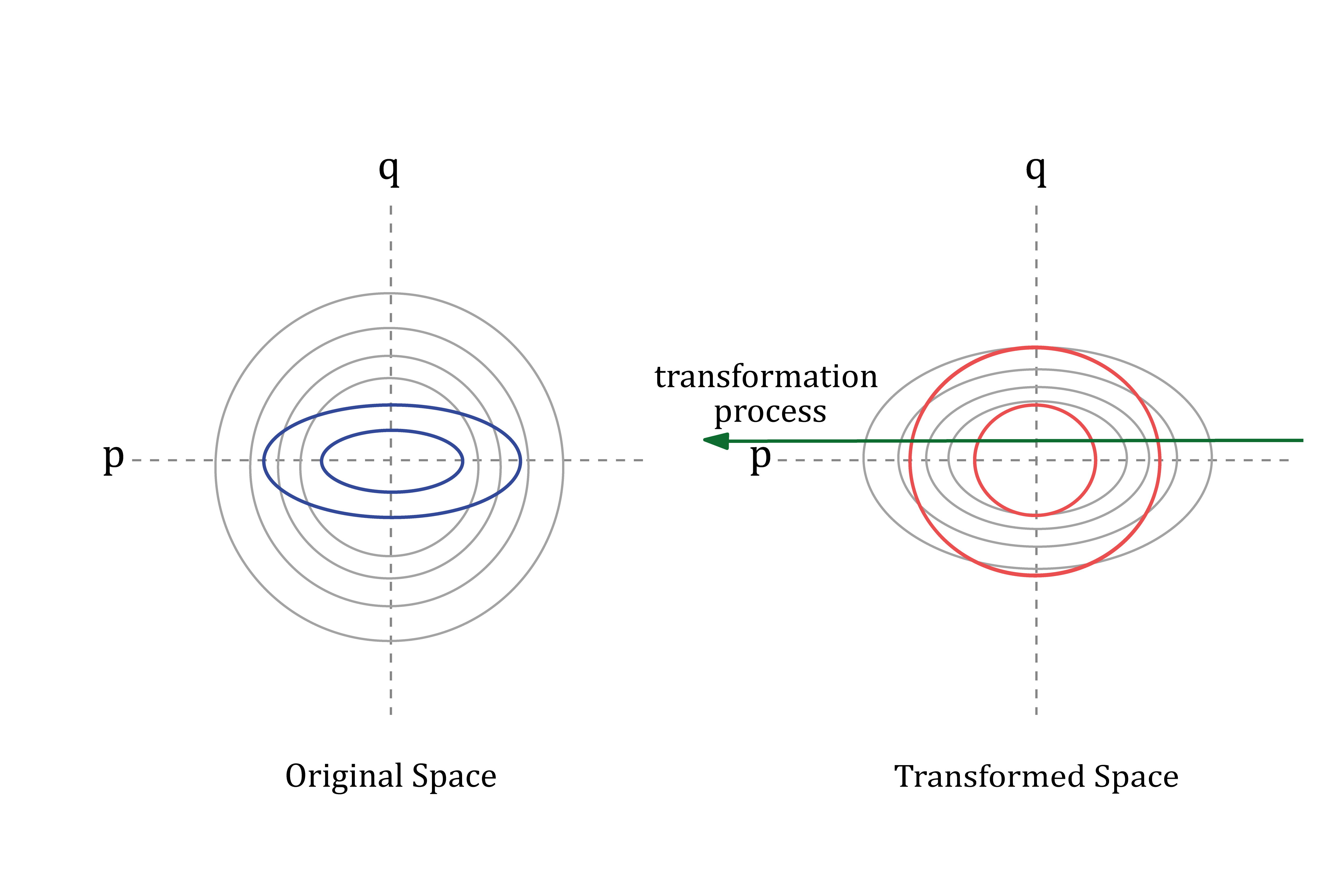}
\caption{Canonical Transformations in Reasoning Space}
\label{fig:canonical-transformations0}
\end{figure}

Figure \ref{fig:canonical-transformations0} illustrates the basic idea of canonical transformations used in our Hamiltonian model. The composition has two phase plots: the ``Original Space'' on the left and the ``Transformed Space'' on the right. The original space denotes the initial model. Elliptical trajectories indicate fluctuating rates of change in states and momentum, while contour lines denote energy levels. The transformed framework illustrates the identical system after a canonical transformation. Circular trajectories indicate a more consistent advancement across states. Modified contour lines indicate a revised energy distribution. $\Phi(x,t)$ correlates original reasoning states with new ones. $H(x,t) + U(x,t)$ modifies the Hamiltonian (total energy) by incorporating an additional term $U(x,t)$. Finally, $y + a(x,t)$ transforms the momentum. The green arrow in the illustration indicates the operation of the canonical transformation. The arrow represents the transformation process from the  original space (left plot) to the transformed space (right plot). It illustrates the mapping of locations, trajectories, and energy levels from the original space to their corresponding elements in the transformed space. The arrow's smoothness indicates that the transformation is continuous and clearly specified for all points within the reasoning space.

\section{A new framework for reasoning systems}
\subsection{Hamiltonian framework for reasoning}
\subsubsection{Defining the reasoning state space}

In our correlation, we can represent reasoning states as vectors in a high-dimensional embedding space \cite{mikolov2013b}. This embedding space is derived from a pre-trained language model \cite{devlin2018}, capturing the semantic content of each reasoning step. Formally, we define the reasoning state $q$ as:

\begin{equation}
q = E(x) \in \mathbb{R}^d
\end{equation}
where $E: V \to \mathbb{R}^d$ is the embedding function, $V$ is the vocabulary of the language model (corpus), and $x \in V^*$ is the input text (e.g., a fact or question in the reasoning chain). The input text $x$ is first tokenized into a sequence of tokens $(t_1, t_2, \ldots, t_n)$ using a tokenizer $T$. Each token $t_i$ is mapped to its corresponding embedding vector $e_i$:

\begin{equation}
e_i = E_{token}(t_i) \in \mathbb{R}^d
\end{equation}

The model processes these token embeddings \cite{sennrich2015} through its layers (e.g., transformer layers) to produce contextual embeddings:

\begin{equation}
(c_1, c_2, \ldots, c_n) = model([e_1, e_2, \ldots, e_n])
\end{equation}

where $c_i \in \mathbb{R}^d$ are the contextual embeddings. Finally, we aggregate these contextual embeddings to represent the entire input:

\begin{equation}
q = A(c_1, c_2, \ldots, c_n)
\end{equation}
where $A$ is an aggregation function \cite{reimers2019}, which could be mean pooling, 

\begin{equation}
q = \left(\frac{1}{n}\right) \sum_i c_i
\end{equation}

max pooling, $q_j = \max_i(c_i)_j$ for each dimension $j$, or $[CLS]$ token: $q = c_1$ (assuming the first token is a special [CLS] token).

\subsubsection{Hamiltonian for reasoning chains}

A reasoning chain can be represented as a sequence of states $Q = (q_1, q_2, \ldots, q_m)$ where each $q_i \in \mathbb{R}^d$. We define a Hamiltonian for reasoning $H_R: \mathbb{R}^d \times \mathbb{R}^d \to \mathbb{R}$ as:

\begin{equation}
H_R(q, p) = T(p) - V(q)
\end{equation}

where $q$ represents the current state of reasoning, analogous to position in mechanical systems, and $p$ represents the change in reasoning, corresponding to momentum \cite{friston2010}. The reasoning momentum $p$ can be defined as the difference between consecutive states $p_i = q_{(i+1)} - q_i$. $T(p)$ is the ``kinetic'' term representing the cost of changing the reasoning state, and $V(q)$ is the ``potential'' term representing the relevance or correctness of the current reasoning state. The notation $R$ denotes a reasoning system. Note that this equation is similar to Lagrangian equation already introduced. Although Lagrangian mechanics is contained in Hamiltonian mechanics as a special case, ``the Hamiltonian point of view allows us to solve completely a series of mechanical problems which do not yield solutions by other means'' \cite{glattfelder2019}. The kinetic term $T(p)$ can be understood as the cognitive effort or the computational cost associated with changing the reasoning state \cite{friston2010}. We define this effort as:

\begin{equation}
T(p) = \frac{1}{2} \|p\|^2
\end{equation}
where $\|p\|$ is the magnitude of the change vector. This quadratic form is analogous to kinetic energy in classical physics, penalizing large and sudden changes in reasoning. The term $V(q)$ represents the degree to which the current reasoning state corresponds with the target question being addressed. A lower potential energy indicates a more relevant or realistic state. We could define it as:

\begin{equation}
V(q) = - sim(q, q_d)
\end{equation}

where $sim(\cdot, \cdot)$ is a similarity function like cosine similarity \cite{rahutomo2012}, and $q_d$ is the embedding of the desired answer or goal state.

\begin{equation}
sim(\vec{q}, \vec{q_d}) = \frac{\vec{q} \cdot \vec{q_d}}{\|\vec{q}\| \|\vec{q_d}\|}
\end{equation}

$\vec{q}$ and $\vec{q_d}$ are represented as two non-zero vectors in an inner product space. The reasoning phase space $(q, p)$ inherits the symplectic structure discussed earlier. This implies that our reasoning Hamiltonian will preserve certain geometric properties as it evolves, analogous to the conservation of phase space volume in dynamical systems \cite{strogatz2018}. Then, we can apply canonical transformations to our reasoning Hamiltonian $H_R(q, p)$, allowing us to change variables while preserving the fundamental structure of the system \cite{arnold2013}. We can write this transformation as:

\begin{equation}
\bar{x} = \Phi(x,t)
\end{equation}
\begin{equation}
\bar{H_R} = H_R(x,t) + U(x,t)
\end{equation}
\begin{equation}
\bar{y} = y + a(x,t)
\end{equation}
where $x$ represents our original phase space variables $(q, p)$, and $(\bar{x}, \bar{y})$ represents the transformed variables. $\bar{H_R}$ is the transformed Hamiltonian, and $U(x,t)$ is a generating function for the transformation.

A special case in Hamiltonian systems is when we represent them in a two dimensional space ($H_R: \mathbb{R}^2 \times \mathbb{R}^2 \to \mathbb{R}$). The importance of recognizing a system as Hamiltonian lies in the ability to build the phase view without requiring a solution to the system \cite{hirsch2013}. Assuming that $H$ is not constant on any open set, we proceed with drawing the level curves $H(q, p) = constant$. Solutions lie on these level sets, with trajectory orientations determined directly by the vector field. It is important to note that the equilibrium points of a Hamiltonian system are located at the critical points of $H$, namely at the places where both partial derivatives of $H$ equal zero.

\begin{figure}[h]
\centering
\includegraphics[width=0.75\textwidth]{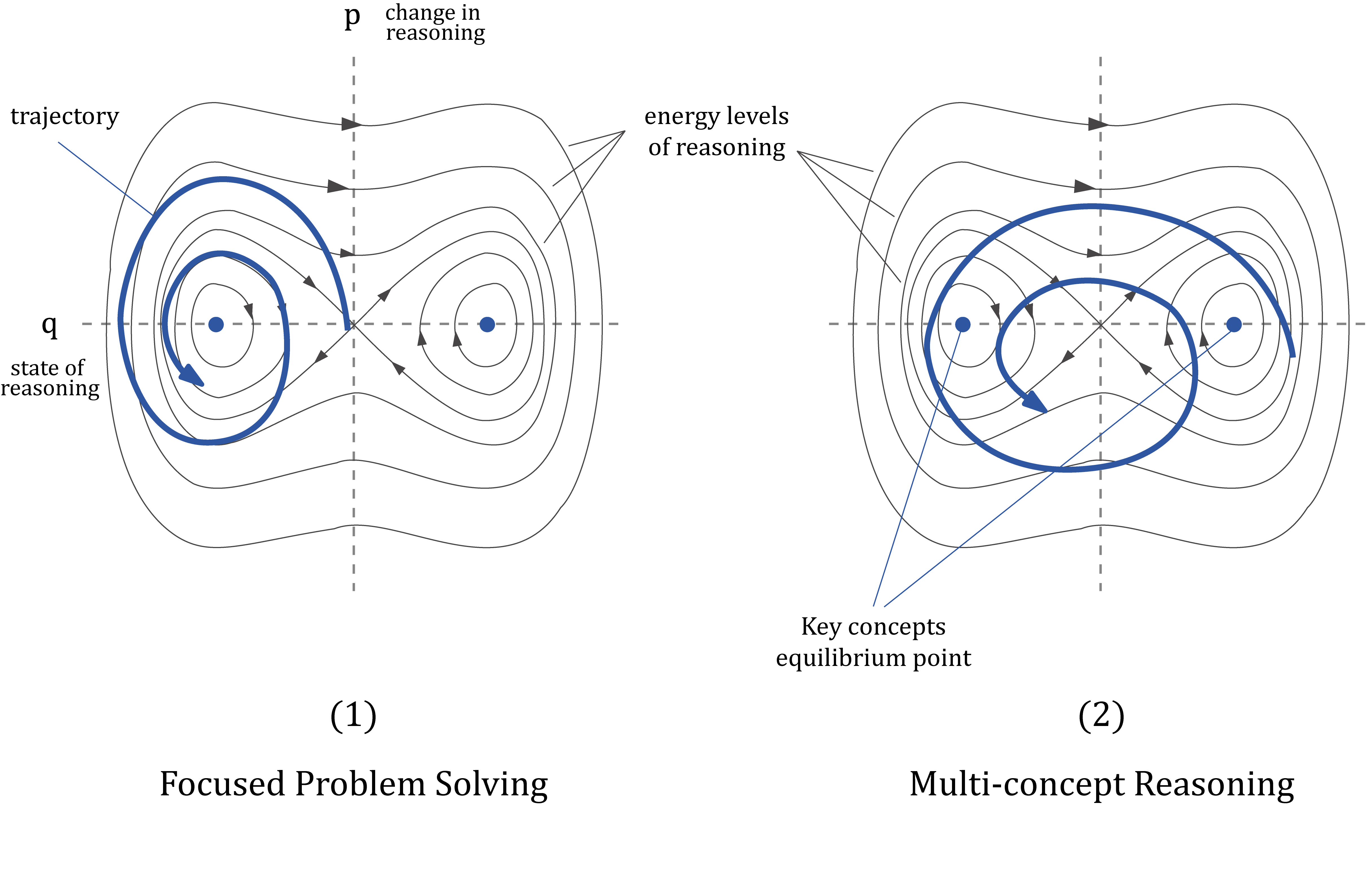}
\caption{Phase plots for focused and multi-concept reasoning in a two-dimensional Hamiltonian system}
\label{fig:phase-plots}
\end{figure}

Figure \ref{fig:phase-plots} illustrates a phase plot for a reasoning system, with the $q$-axis denoting the current state of reasoning, similar to position in mechanical systems, and the $p$-axis indicating the change in reasoning, analogous to momentum. The contour lines denote the energy levels associated with reasoning. The blue lines illustrate potential reasoning paths and the evolution of reasoning over time within the phase space. The blue dots indicate stable states or key concepts. Figure 2.1 illustrates a focused problem-solving approach, characterized by tighter orbits around a singular concept, reflecting a concentrated emphasis on a single concept. Figure 2.2 illustrates a multi-concept reasoning framework, illustrating a larger orbit that includes several key concepts, thereby representing the integration of multiple reasoning levels.

In the context of our reasoning system, these transformations allow us to change the representation of our reasoning state while preserving the fundamental structure of the system. It also introduces new variables that may provide insights into the reasoning process, simplifying the analysis of the reasoning dynamics by choosing appropriate transformations. For example, we can use a transformation to move from a word embedding space to a more abstract concept space, or to focus on particular aspects of the reasoning process. The ability to perform these transformations, while maintaining the canonical structure of our Hamiltonian, is key for the flexibility and capability of this approach. Then, we can analyze the reasoning process from multiple perspectives and at different levels of abstraction, all within the same theoretical framework. The geometry of this embedding space is important for understanding reasoning dynamics \cite{pennington2014} and will be presented in section 3.2.

\subsubsection{Calculation of Hamiltonian energies for reasoning}

We assume an optimal reasoning process in which the total energy $H_R$ remains invariant. This implies a trade-off between exploration (high $T$, low $V$) and exploitation (low $T$, high $V$) during the reasoning process. This trade-off is analogous to key principles in reinforcement learning and statistical physics. In RL, it proves as the balance between trying new actions (exploration) and leveraging known good strategies (exploitation) \cite{kaelbling1996}. In physical systems, it appears in phenomena like simulated annealing \cite{bertsimas1993}, where high temperatures allow broad exploration of state space, while low temperatures exploit known low-energy configurations. Our Hamiltonian formulation provides a rigorous mathematical framework for analyzing this trade-off in reasoning processes, potentially leading to new insights into optimal reasoning strategies and their connections to learning and physical systems. A practical application of this trade-off in reasoning processes would be to explicitly define $T(p)$ and $V(q)$ in terms of exploration and exploitation measures in our reasoning space. Therefore, we can analyze how different reasoning strategies balance $T$ and $V$ over time to find successful reasoning chains with a particular $T/V$ ratio or evolution pattern. We can explore associations with Reinforcement Learning algorithms, potentially adapting approaches such as Thompson sampling \cite{russo2018}, or intrinsic motivation \cite{barto2013}, to guide reasoning processes. It will be very stimulating to consider quantum analogies, where ``superposition'' could represent simultaneous exploration of multiple reasoning paths \cite{elsage2022}.

To calculate the Hamiltonian energies for each reasoning chain, we follow these steps:
\begin{enumerate}[label=\alph*)]
\item Embed each fact and question in the reasoning chain using the embedding function $E$.
\item Calculate $p = q_{t+1} - q_t$ as the difference between consecutive reasoning states.
\item Compute $T(p)$ and $V(q)$.
\item Calculate the total Hamiltonian energy $H_R$.
\end{enumerate}

We perform these calculations for each step in the reasoning chain, allowing us to analyze the energy profile of the complete reasoning process.

\subsection{Geometric analysis of reasoning trajectories}

The application of differential geometry to reasoning trajectories offers a robust framework for studying the structure and attributes of cognitive processes. By conceptualizing reasoning paths as curves within a high-dimensional space, we can apply mathematical tools from differential geometry to measure and describe the properties of these paths \cite{amari2016}. With this approach we can move beyond simple distance metrics in embedding spaces and consider the intrinsic geometry of reasoning trajectories.

\subsubsection{Differential geometry in cognitive spaces}

As we have already introduced, our framework considers reasoning processes as paths $\gamma(t)$ in a high-dimensional manifold $M$, which represents the space of possible cognitive states. This manifold is provided with a metric $g$, which defines distances and angles in the cognitive space \cite{do2016}. The metric captures the semantic similarity between different cognitive states and can be derived from embedding models such as BERT or GPT \cite{devlin2018}. The tangent vector $\gamma'(t)$ at each point represents the immediate direction and velocity of reasoning, while higher-order derivatives capture how this direction changes over time. With this geometric approach we can analyze both the content and the dynamics of reasoning processes.

\subsubsection{Trajectories' curvature and cognitive flexibility}

One of the key geometric properties we can analyze is the curvature of reasoning trajectories. For a curve $\gamma(t)$, the curvature $\kappa$ at a point is given by:

\begin{equation}
\kappa = \frac{|\gamma''(t) \times \gamma'(t)|}{|\gamma'(t)|^3}
\end{equation}
where $\times$ denotes the cross product and $|\cdot|$ the magnitude \cite{oneill2006}. In the context of reasoning, curvature can be interpreted as a measure of ``cognitive flexibility'' or the rate at which the direction of reasoning chain evolves. High curvature indicates rapid shifts in reasoning direction, potentially representing moments of insight or even the integration of diverse ideas. Low curvature suggests more linear, focused reasoning \cite{sussillo2013}.

\subsubsection{Frenet-Serret framework and multi-aspect reasoning}

\begin{figure}[h]
\centering
\includegraphics[width=0.75\textwidth]{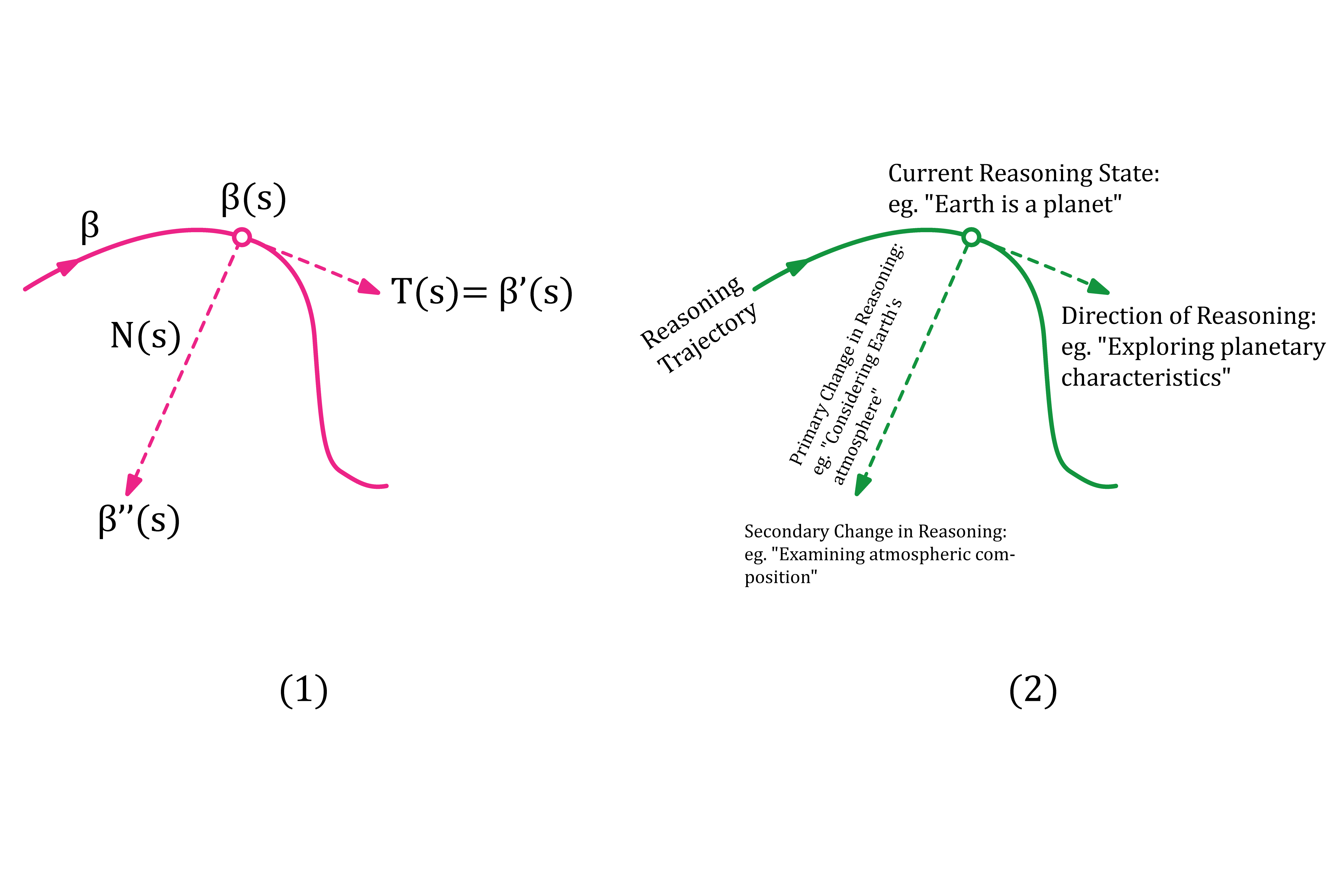}
\caption{Representation of curvature with Frenet frame field.}
\label{fig:frenet-frame}
\end{figure}

The Frenet-Serret theorems provide mathematical measurements for turning and twisting a curve in $\mathbb{R}^3$. Let $\beta: I \longrightarrow \mathbb{R}^3$ be a unit speed curve with curvature $\kappa > 0$ and torsion $\tau$ \cite{oneill2006}

\begin{equation}
T'(t) = \kappa(t)N(t)
\end{equation}
\begin{equation}
N'(t) = -\kappa(t)T(t) + \tau(t)B(t)
\end{equation}
\begin{equation}
B'(t) = -\tau(t)N(t)
\end{equation}
$T$, $N$, and $B$ are the tangent, normal, and bi-normal unit vectors. $T = \beta'$ is the unit tangent vector field of $\beta$, and has a constant length of 1. Thus, its derivative $T' = \beta''$ measures how the curve is turning. $N$ is the principal vector field of $\beta$, and $B = T \times N$ is the bi-normal vector field of $\beta$ (figure \ref{fig:frenet-frame}).

In our framework, $T$ represents the current direction of reasoning, $N$ indicates the primary direction of change in reasoning, $B$ captures secondary changes orthogonal to both $T$ and $N$, and $\tau$ quantifies how the osculating plane (spanned by $T$ and $N$) changes along the curve (Figure \ref{fig:frenet-frame}). Consequently, we can analyze not just the ``bendiness'' of a reasoning path, but also how it twists in the high-dimensional concept space, providing insights into multi-aspect reasoning processes.

Figure 2.1 illustrates the progression of a reasoning chain from an initial point, altering its trajectory as new elements are evaluated, potentially diverging into secondary issues, all while preserving the geometric connections defined by the Frenet-Serret framework. The Frenet frame offers a more natural method to visualize and understand complex reasoning processes in LLMs. High curvature or torsion points in reasoning processes may represent critical choice points or insights, facilitating targeted interventions or optimizations. Different reasoning processes may be geometrically contrasted, thereby facilitating the identification of more efficient or successful methods.

\begin{figure}[h]
\centering
\includegraphics[width=0.75\textwidth]{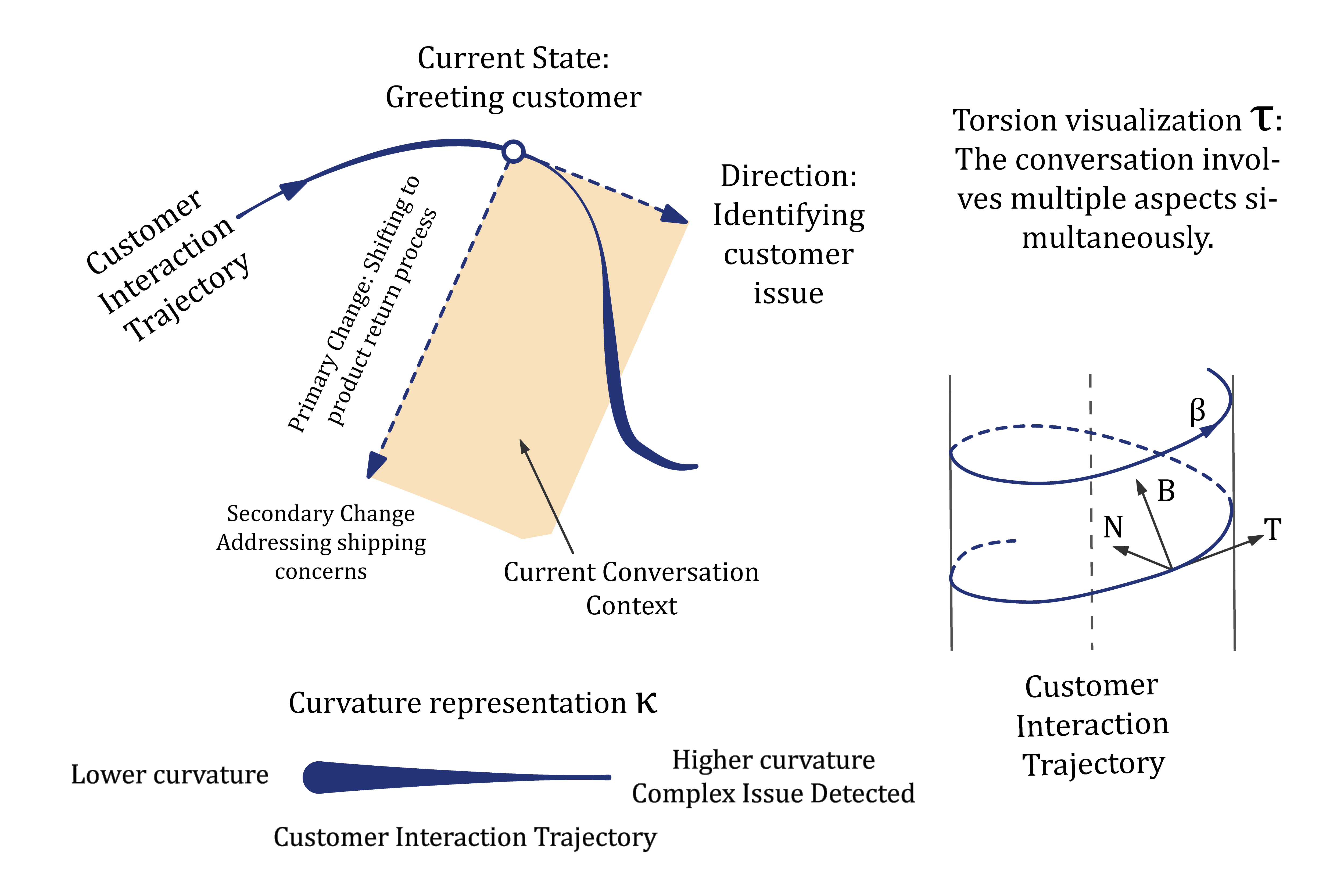}
\caption{Representation of curvature for a large e-commerce company chatbot example.}
\label{fig:chatbot-example}
\end{figure}

Consider a large e-commerce company deploying an AI-driven customer attention chat-bot. With this framework to analyze the chat-bot's reasoning processes, the organization could identofy client interactions, with each interaction represented as a curve within the reasoning space (Figure \ref{fig:chatbot-example}). We can get insights about the current direction of reasoning, $T(s)$, identify shifts in topic during conversation, $N(s)$, or unforeseen situations during the conversation, $B(s)$. The curvature, $\kappa$, would show the rate at which the discussion is altering direction, while the torsion, $\tau$, would indicate the conversational framework is evolving (Figure \ref{fig:chatbot-example}). Geometric analysis enables identifying optimal low-curvature trajectories for common inquiries, while recognizing when high curvature is necessary for complex problems requiring strategic shifts.

Torsion analysis can be used to prepare the chat-bot for complex, multi-concept challenges (Figure \ref{fig:phase-plots}). In summary, Frenet frames enable assessment of successful interactions, identification of ideal geometric patterns, and implementation of real-time dialogue analysis to dynamically adapt chat-bot strategies.

\subsubsection{Arbitrary speed curves}

Given a regular curve $\beta: I \to \mathbb{R}^3$ with speed function $v$, we can calculate its velocity and acceleration \cite{oneill2006} as:

\begin{equation}
\beta' = vT
\end{equation}
\begin{equation}
\beta'' = \frac{dv}{dt}T + \kappa v^2 N
\end{equation}
where $T$ is the tangent to the curve, and $\kappa$ is the curvature in Frenet frame. This equation has two parts: $\frac{dv}{dt}T$ is a tangential component and measures the rate of change of $\beta'$, and $\kappa v^2 N$ is the normal that points perpendicular to motion. The normal is a vector that represents a force acting at a 90-degree angle to the axis of motion, which does not alter the object's velocity in the direction of its original motion. According to Newton's laws of motion, this normal will create an angle such that a part of its velocity is now aligned with the normal direction. The two components combine to generate diagonal velocity at an angle dependent upon the magnitude of the applied force.

\begin{figure}[h]
\centering
\includegraphics[width=0.75\textwidth]{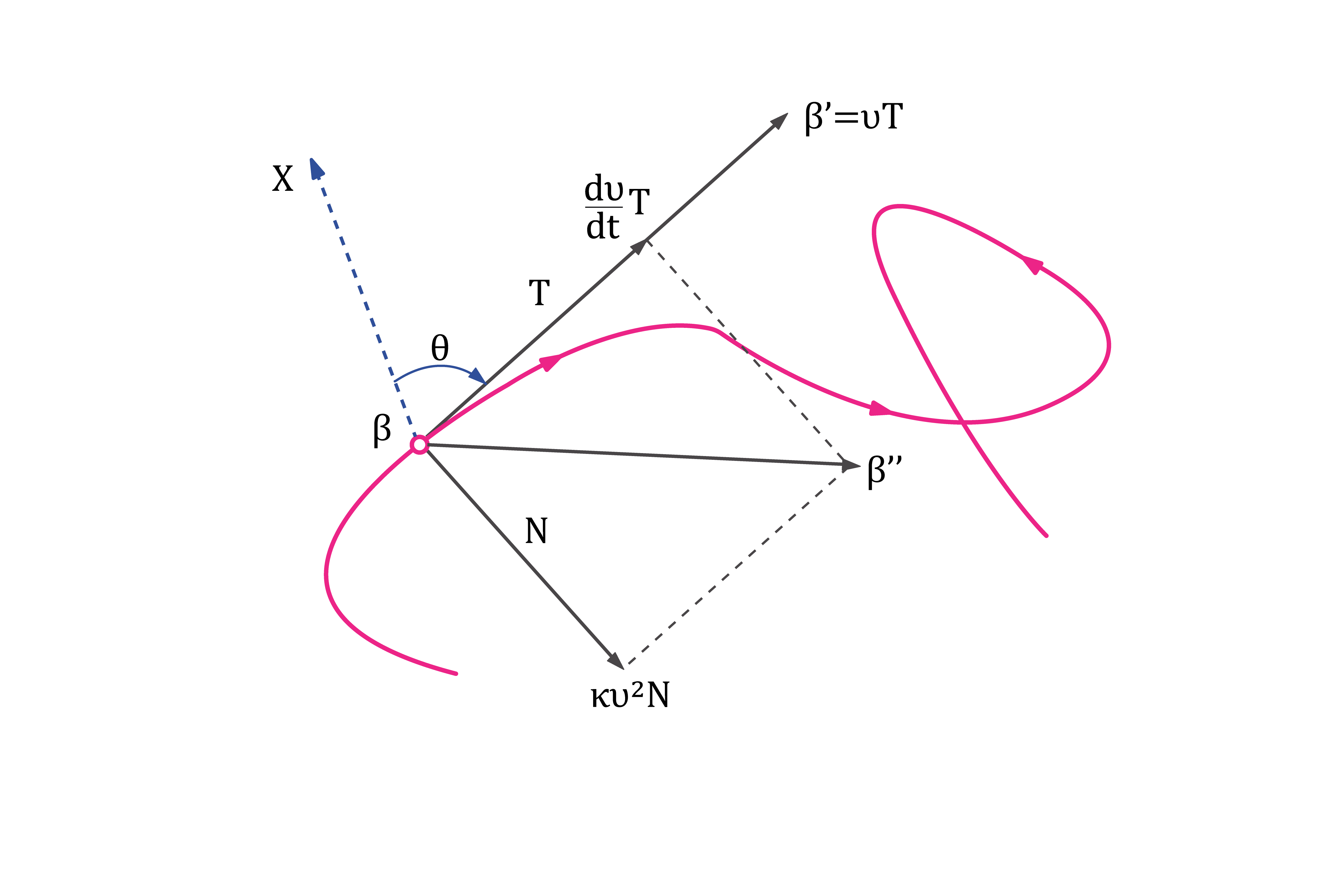} 
\caption{Velocity, acceleration and trajectory angle in a curve using Frenet frame.}
\label{fig:velocity-acceleration}
\end{figure}

In the context of reasoning, the velocity of reasoning defines a ``magnitude''. A higher magnitude implies swifter transitions between ideas, whereas a smaller magnitude denotes a more gradual progression. This magnitude likewise represents the ``distance'' between successive concepts in reasoning. Large magnitude values represent substantial leaps between divergent ideas, whereas small magnitude values represent a progressive evolution between closely related ideas. Acceleration in reasoning represents the changes in velocity over time. An increase in acceleration suggests an intense rate of ideation in problem-solving, while a decrease in acceleration indicates a deliberate deceleration to concentrate on particular elements.

In Frenet frame, the trajectory angle $\theta$ can be seen as the angle between the tangent vector $T$ and a fixed reference direction (Figure \ref{fig:velocity-acceleration}). This angle changes as the curve evolves, reflecting changes in reasoning process direction. The rate of change $d\theta$ is  correlated with the curvature $\kappa$. The term $\kappa v^2 N$ represents the normal component of acceleration, which is responsible for changing the direction of velocity. We can define the trajectory angle $\theta$ as

\begin{equation}
\frac{d\theta}{dt} = \kappa v
\end{equation}

This equation links the angular rate of change with the trajectory's curvature, velocity and magnitude. A high magnitude could indicate creative, divergent thinking processes \cite{beaty2018}, while decreasing magnitude over time could represent a convergence towards a solution or conclusion \cite{zabelina2016}. Sudden large magnitude transitions could correspond to moments of insight or breakthrough in problem-solving \cite{kounios2014a}, while consistent, moderate magnitude transitions could indicate systematic, analytical thinking \cite{helie2010}. Understanding these dynamics can help in analyzing and potentially optimizing reasoning processes in both human cognition and language models. A model fine-tuned for creative work could be optimized for rapid transitions, whereas one built for analytical tasks can be tuned for steady, measured progressions. Cognitive processes can be viewed as continuous trajectories in state spaces \cite{spivey2006}. The trajectory angle $\theta$ could represent the current direction of reasoning in the conceptual space. Changes in $\theta$ indicate shifts in the focus or approach of the reasoning process. For example, sudden changes in $\theta$ might correspond to creative leaps or relevant insights \cite{beaty2014},meanwhile slow, steady changes in $\theta$ could represent methodical and analytical reasoning \cite{kounios2014b}. Magnitude represents the rate of concept progression through the model, whereas $\theta$ denotes the trajectory. In combination, they offer a more comprehensive understanding of the trajectory of reasoning.

\subsection{Symmetry and conservation laws in reasoning processes}

 The connection between symmetry and conservation, first formalized by Emmy Noether in 1918 \cite{kosmann2011}, reveals deep insights into the nature of invariances in physical systems. Our analysis of reasoning trajectories through the lens of Hamiltonian mechanics and Lie group theory highlights how conserved quantities emerge from system symmetries \cite{glattfelder2019}. In classical physics, the invariance of physical laws under time translations leads to energy conservation, just as we find analogous conserved quantities in our reasoning space.

A symmetry can be described as a transition that preserves certain properties of a system \cite{glattfelder2019}. For a simple example, consider rotating a perfect circle 90º; it remains the same afterward. Mathematically, a group action on a set $X$ is defined as a function

\begin{equation}
\Phi: G \times X \to X
\end{equation}

and is a symmetry group if its group action $\Phi$ preserves the structure on $X$, that is, leaves $X$ invariant. For example, a square $(X)$ has rotational symmetry $(G)$ - it looks the same after rotating 90 degrees. But the concept of symmetry expanded beyond geometric shapes. It became about transformations that preserved particular properties, even in more complex abstract spaces like embedding spaces. Many natural laws are symmetrical. For example, the laws of physics apply the same sense regardless of where we are in space (implying translational symmetry), or the direction we take (rotational symmetry). A Lie group $G$ is a continuous transformation group that is also a differentiable manifold \cite{arnold2013}, with the group operations being differentiable maps. A symmetry $S$ can be defined as

\begin{equation}
S(t) = exp(tX^a)
\end{equation}

where $S(t) \in G$, $t \in \mathbb{R}$ and $X^a$ is called generator. Lie groups give a language for describing and analyzing continuous symmetries with precision \cite{duistermaat2012}.

As said, Noether's theorem \cite{kosmann2011} defines an important connection between symmetries and conservation laws in physics. For any continuous symmetry in a physical system, there exists a corresponding conserved quantity that remains invariant across time. A symmetry implies an invariant system property under a transformation. For instance, running an experiment today or tomorrow (temporal translation) will provide consistent physical laws. A conserved quantity is the property that remains unchanged while a system evolves. The invariance of physical theories about spatial and temporal transformations results in conservation laws, including the conservation of momentum and energy in the universe. Noether's theorem addresses the deep connection in nature: the symmetries observed in the world are closely linked to the conserved quantities in the transformation of physical systems. In other words, for each symmetry identified in an embedding space, there must be a corresponding conserved quantity in this space. The canonical transformations we applied to our reasoning trajectories, transforming from the original phase space to action-angle variables, unveil that while the ``energy'' (action) of reasoning processes tends to be conserved, the ``phase'' (angle) varies. This mirrors the behavior of classical mechanical systems and suggests that effective reasoning maintains a consistent level of complexity or engagement while exploring different cognitive directions. Our approach uses a Hamiltonian $H_R(q, p)$ for reasoning, where $q$ represents the current state of reasoning and $p$ the change in reasoning. This Hamiltonian system is analogous to those in classical mechanics, but is applied to abstract reasoning spaces. The Hamiltonian evolution along reasoning trajectories can be represented as:

\begin{equation}
\frac{dq}{dt} = \frac{\partial H_R}{\partial p}
\end{equation}
\begin{equation}
\frac{dp}{dt} = -\frac{\partial H_R}{\partial q}
\end{equation}

It provides a mathematical description of how reasoning processes unfold in our abstract space. The analysis of trajectory properties, such as curvature $\kappa$ and torsion $\tau$, provides quantitative measures of symmetries in reasoning patterns, offering insights into the ``cognitive flexibility'' of reasoning processes.

\begin{equation}
\kappa = \frac{\|T'\|}{\|T'\|}
\end{equation}
\begin{equation}
\tau = \frac{-B \cdot (N N')}{\|N \times N'\|}
\end{equation}

where $T$, $N$, and $B$ are the tangent, normal, and bi-normal vectors of the Frenet-Serret frame.

\section{Methodology}
\subsection{Dataset description}

We used the OpenBookQA (OBQA) dataset for our research, which provides a standard to assess the question answering and reasoning abilities of AI systems. The OBQA dataset was presented by Mihaylov et al. \cite{mihaylov2018} in their research on open-book question answering. It was developed to evaluate AI systems' capacity to respond to inquiries necessitating the integration of information from a specified text corpus with general knowledge. The dataset emulates open-book examinations, wherein a limited collection of fundamental facts is supplied, requiring the integration of this information with general knowledge to respond to questions. The dataset centers on elementary science topics, being appropriate for assessing factual memory and reasoning skills. The dataset has been built to create difficulties for contemporary foundation models (synthesis of retrieval, reasoning, and common sense comprehension).

The OBQA dataset contains 5,957 multiple-choice questions (4,957 for training, 500 for testing), each with four options and one correct answer. Questions require multi-step reasoning that combines provided facts with general knowledge. Unlike other datasets, OBQA doesn't include explanations or reasoning chains, making it ideal for testing explanation generation models. Questions span various reasoning types including causation, purpose, and property attribution. Our experiment focuses on generating and annotating reasoning chains for the 500 test questions using the QASC corpus and our proposed technique.

\subsection{Implementing the Hamiltonian framework for NLP}
 
\subsubsection{Embedding representation}

We use a BERT-based model to analyze and build reasoning chains. BERT (Bidirectional Encoder Representations from Transformers), developed by Devlin et al. \cite{devlin2018}, is a transformer-based methodology for natural language processing. We selected BERT due to its exceptional performance in several NLP tasks, such as question answering and natural language inference. Our BERT-based model is optimized for the task of recognizing valid reasoning chains. The system accepts a question, an answer, and a proposed reasoning chain, subsequently producing a score that reflects the chain's validity. The architecture of the model comprises a BERT-base-uncased model serves as the primary encoder and a specialized layer above BERT for binary classification (valid/invalid chain). Input formatting that amalgamates the question, answer, and reasoning chain sentences, delineated by [SEP] tokens. We incorporate the OBQA dataset with our BERT-based model and the QASC corpus to create a framework that enables both question answering and the generation and assessment of explanations for those answers, thereby addressing the problem of explainable AI in multi-hop reasoning tasks.

\subsubsection{Basic concepts}

Our implementation allows us to apply Hamiltonian framework to discrete linguistic elements. Each step in a reasoning chain is treated as a discrete point in our continuous embedding space, forming a trajectory that can be analyzed using our framework. We can define the basic elements in our Hamiltonian framework as follows:

\begin{enumerate}[label=\alph*)]
\item Position ($q$): Represented by the BERT embedding of each fact or question in the reasoning chain.
\item Momentum ($p$): Calculated as the difference between consecutive embeddings in the chain.
\item Kinetic Energy ($T$): Defined as the squared magnitude of the momentum, representing the ``cost'' of transitioning between reasoning states.
\item Potential Energy ($V$): Computed using the cosine similarity between the current state and the question embedding, representing the relevance of the current reasoning step to the overall question.
\item Hamiltonian Energy ($H_R$): Calculated as $T - V$, balancing the progression of reasoning against its relevance.
\end{enumerate}

\subsection{Analytical approaches}

\subsubsection{Energy analysis}

We aim to distinguish the distribution of Hamiltonian energy across valid and invalid reasoning chains. This analysis is based on the principle that the Hamiltonian, $H_R = T - V$, represents the total ``energy'' of a reasoning process \cite{goldstein2002}. We use several statistical measures and visualizations to identify patterns that represent effective reasoning.

\subsubsection{Trajectory analysis}

We apply Principal Component Analysis (PCA) to reduce the dimensionality of BERT embeddings, allowing for both visualization and analysis of reasoning trajectories \cite{jolliffe2016}. We formalize these trajectories using geometric properties derived from differential geometry \cite{do2016}:

\begin{enumerate}[label=\alph*)]
\item Magnitude, $v$: Representing the ``velocity'' of cognitive advancement.
\item Angle, $\theta$: Indicating changes in reasoning direction.
\item Curvature, $\kappa$: Quantifying the rate of change in trajectory direction.
\item Torsion, $\tau$: Measuring how the trajectory twists in three-dimensional space.
\end{enumerate}

These properties provide insights into the dynamics of reasoning processes and how they differ between valid and invalid chains.

\subsubsection{Conservation laws}

Using the basic idea of Noether's theorem \cite{kosmann2011}, we explore whether quantities analogous to conserved physical quantities emerge in our reasoning trajectories. We analyze the conservation of certain combinations of trajectory properties across reasoning steps, which could indicate underlying symmetries in the reasoning process.

\subsubsection{Canonical transformations}

We explore alternative representations of reasoning dynamics through transformations inspired by classical mechanics \cite{arnold2013}. By mapping our original phase space $(q, p)$ to new coordinates (e.g., action-angle variables), we aim to unveil hidden patterns and invariants in the reasoning process.

\subsection{Experimental setup}

The analysis of the reasoning chains uses a synthesis of natural language processing methodologies together with Hamiltonian-inspired calculations. Our methodology primarily involves utilizing BERT (Bidirectional Encoder Representations from Transformers) to generate significant representations of reasoning states.

\begin{enumerate}[label=\alph*)]
\item We use a pre-trained BERT model (bert-base-uncased) to produce embeddings for each element of the reasoning chains \cite{devlin2018}. For every fact and inquiry in a reasoning sequence, we derive a high-dimensional vector representation utilizing BERT's last hidden layer. These embeddings encapsulate the semantic essence of each reasoning step in a mathematically analyzable format.
\item We compute Hamiltonian energies for each reasoning chain utilizing BERT embeddings. In our Hamiltonian $H_R$, $q$ denotes the existing state of reasoning (BERT embedding of a fact), and $p$ denotes the variation in reasoning (the difference between consecutive BERT embeddings). The potential term is computed using the cosine similarity between $q$ and the question embedding. We analyze the distribution of Hamiltonian energy in valid versus invalid reasoning chains. We use different statistical analyses and visual representations to identify effective reasoning energy patterns.
\item We apply Principal Component Analysis (PCA) to reduce the dimensionality of the BERT embeddings for the purpose of visualization of the analysis of trajectories in embedding space. We analyze the characteristics of these trajectories, including length, smoothness, curvature, and torsion inside the restricted space. We calculate the geometric characteristics of the trajectories within the BERT embedding space:
   \begin{enumerate}[label=\roman*.]
   \item Magnitude, $v$: Indicating the ``velocity'' of cognitive advancement.
   \item Angle $\theta$: Signifying alterations in the trajectory of reasoning.
   \item Curvature, $\kappa$: Assessing the rate of alteration in the trajectory of reasoning.
   \end{enumerate}
We use different statistical tests to compare these measurements between valid and invalid chains.
\end{enumerate}

\section{Results}

Our BERT-based model, fine-tuned for the goal of recognizing valid reasoning chains, has shown strong performance in discerning between valid and invalid chains. The differentiation between valid and invalid chains can be seen in figure \ref{fig:energy-trajectories}.

\subsection{Hamiltonian framework}

\begin{figure}[h]
\centering
\includegraphics[width=0.75\textwidth]{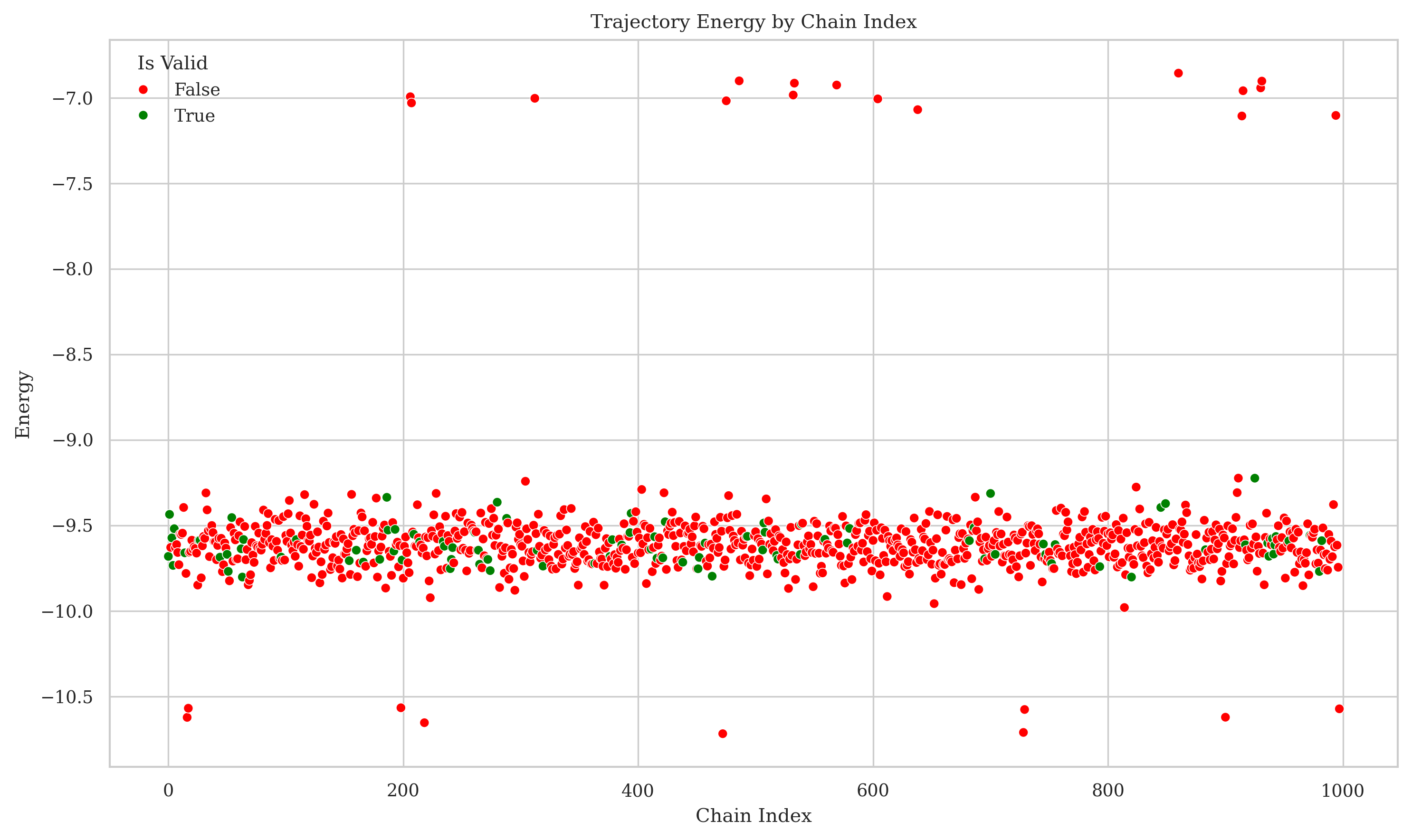} 
\caption{Energies of trajectories for the sample chains within the OBQA dataset. Valid chains (green), invalid chains (red).}
\label{fig:energy-trajectories}
\end{figure}

The main reasoning chains (figure \ref{fig:energy-trajectories}) are grouped within an energy range from -9.5 to -9.0. This implies that most reasoning processes (valid and invalid) operate in a rather small energy range. The distinction in energy levels between valid and invalid chains appears ambiguous, which is remarkable and somewhat paradoxical. The prevalence of high-energy outliers predominantly within invalid chains may suggest that highly ``energetic'' or complex reasoning processes are more susceptible to errors or incorrect results. In contrast, very low-energy outliers correspond mainly to invalid chains, indicating oversimplified or short thinking.

\begin{figure}[h]
\centering
\includegraphics[width=0.75\textwidth]{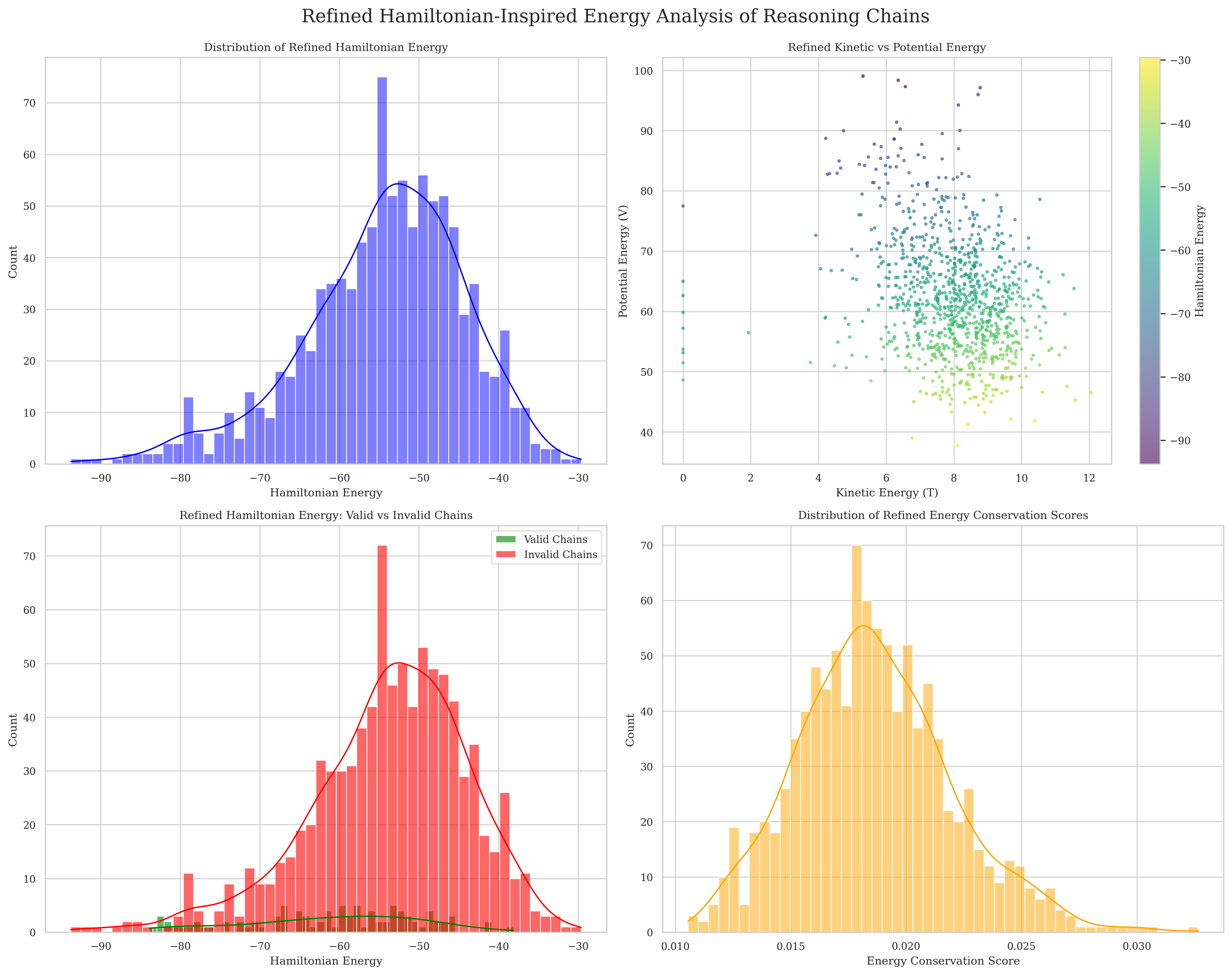} 
\caption{Distribution of Hamiltonian energy in valid and invalid chains within the OBQA dataset. Hamiltonian energy (top left); kinetic vs. potential energy (top right); Hamiltonian energy valid/invalid chains (bottom left); energy conservation score (bottom left).}
\label{fig:hamiltonian-distribution}
\end{figure}

These significant energy states appear to be mostly linked to invalid chains. The generally stable energy band across chain indices indicates that the energy of a reasoning chain is not much influenced by its particular content or length, but rather by intrinsic characteristics of the reasoning process.

\begin{table}[h]
\centering
\begin{tabular}{lc}
\hline
\multicolumn{2}{c}{Distribution of Hamiltonian energies} \\
\hline
Average Energy Conservation Score & 0.019 \\
Correlation between Energy Conservation and Validity & -0.189 \\
\hline
\multicolumn{2}{c}{t-test for difference in Hamiltonian Energy} \\
t-statistic: & -6.5304 \\
p-value: & 0.0001 \\
\hline
\end{tabular}
\caption{Energy conservation scores and Hamiltonian energy}
\label{tab:energy-conservation}
\end{table}

The Hamiltonian-inspired energy framework for analyzing reasoning chains shows several key insights into cognitive processes (Figure \ref{fig:hamiltonian-distribution}). The energy distribution across chains shows a consistent pattern, suggesting a fundamental stability in reasoning regardless of content. There is a clear correlation between kinetic and potential energy, indicating that as reasoning becomes more dynamic, it also tends to involve deeper or more complex concepts.

Valid reasoning chains generally show higher energy levels and more variability compared to invalid chains. Sound logical reasoning often demands deeper cognitive engagement, though our framework reveals multiple valid pathways with distinct energy signatures.

Energy conservation within chains appears to be a common feature, implying that effective reasoning maintains a balance between introducing new ideas and making logical connections. Interestingly, invalid chains mainly show lower energy profiles, which may indicate the use of cognitive shortcuts or simplistic associations.

The framework's ability to differentiate between valid and invalid reasoning chains based on energy levels offers promising applications in assessing LLMs logical processes. Moreover, statistical analysis confirms significant differences in the energy patterns of valid versus invalid chains.

This approach reveals parallels between cognitive and physical systems through shared energetic principles. Viewing reasoning as a physical process offers fresh insights into logical thinking and may unlock new ways to understand cognitive mechanisms.

\begin{figure}[h]
\centering
\includegraphics[width=0.75\textwidth]{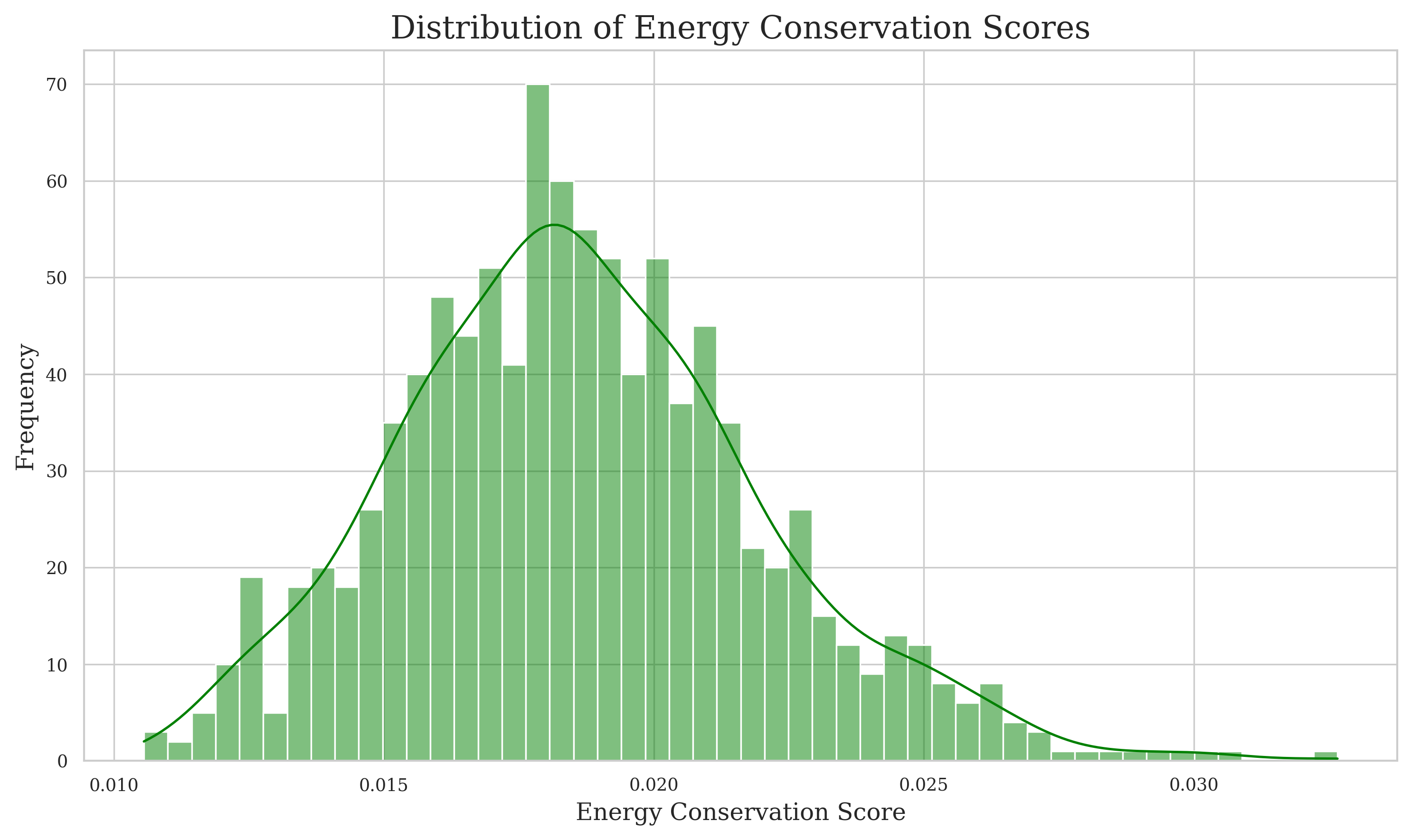} 
\caption{Distribution of energy conservation score for the sample chains within the OBQA dataset.}
\label{fig:energy-conservation}
\end{figure}

Energy Conservation Scores measure the extent to which the total energy of a reasoning chain is maintained along its trajectory. These scores are generally obtained from the balance between kinetic energy (indicating ``movement'' or shifts in reasoning) and potential energy (depth or complexity of the concepts involved). The distribution is approximately normal, with a small right skew. Figure \ref{fig:energy-conservation} illustrates the distribution of Energy Conservation Scores for reasoning chains. Main results range between 0.018 and 0.022, showing a moderate level of energy saving in common reasoning chains. The complete scores range from approximately 0.010 to 0.032, indicating diversity in the energy chain's conservation value. The peak of the distribution indicates the optimal or significant level of energy conservation in cognitive operations. A limited number of chains show either very low or very high conservation scores, potentially indicating irregular or limit reasoning patterns. Within our framework, this distribution suggests that most reasoning chains achieve a balance between energy change and conservation, but certain chains show deeper patterns of energy dynamics throughout the reasoning process.

\begin{figure}[h]
\centering
\includegraphics[width=0.75\textwidth]{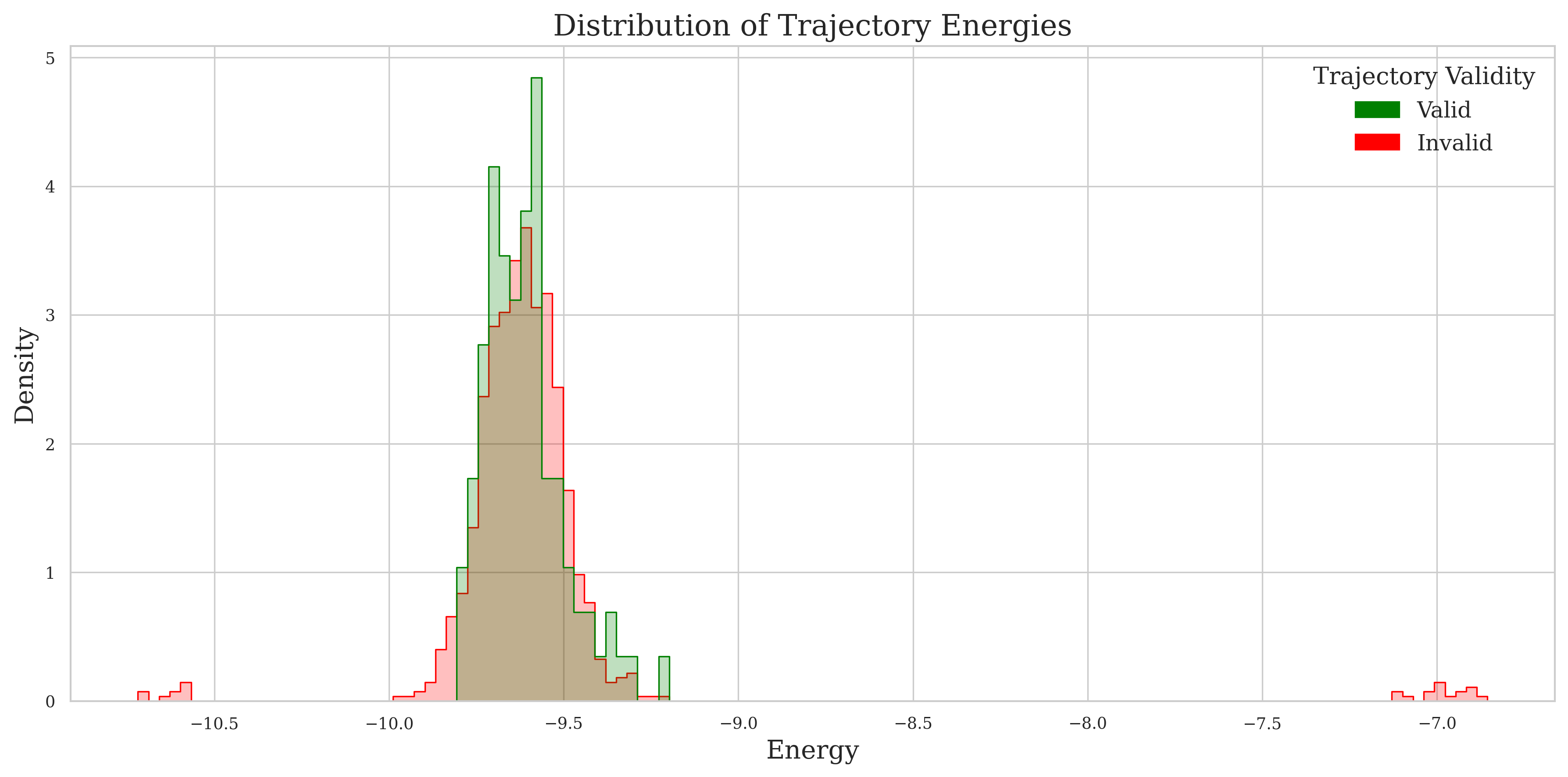} 
\caption{Distribution of trajectory energies for valid (green)and invalid chains (red).}
\label{fig:trajectory-energies}
\end{figure}

Figure \ref{fig:trajectory-energies} shows energy distributions for valid and invalid reasoning chains. Both distributions overlap significantly and center between -9.8 and -9.5, but with key differences. Valid trajectories (green) form a narrower, taller distribution, indicating more consistent energy profiles. Invalid trajectories (red) spread more widely with extended tails at both energy extremes, suggesting that both overly simple (low-energy) and highly complex (high-energy) reasoning can lead to incorrect conclusions. This overlap demonstrates that energy alone cannot definitively determine reasoning validity, highlighting the need for additional assessment criteria.

\begin{figure}[h]
\centering
\includegraphics[width=0.75\textwidth]{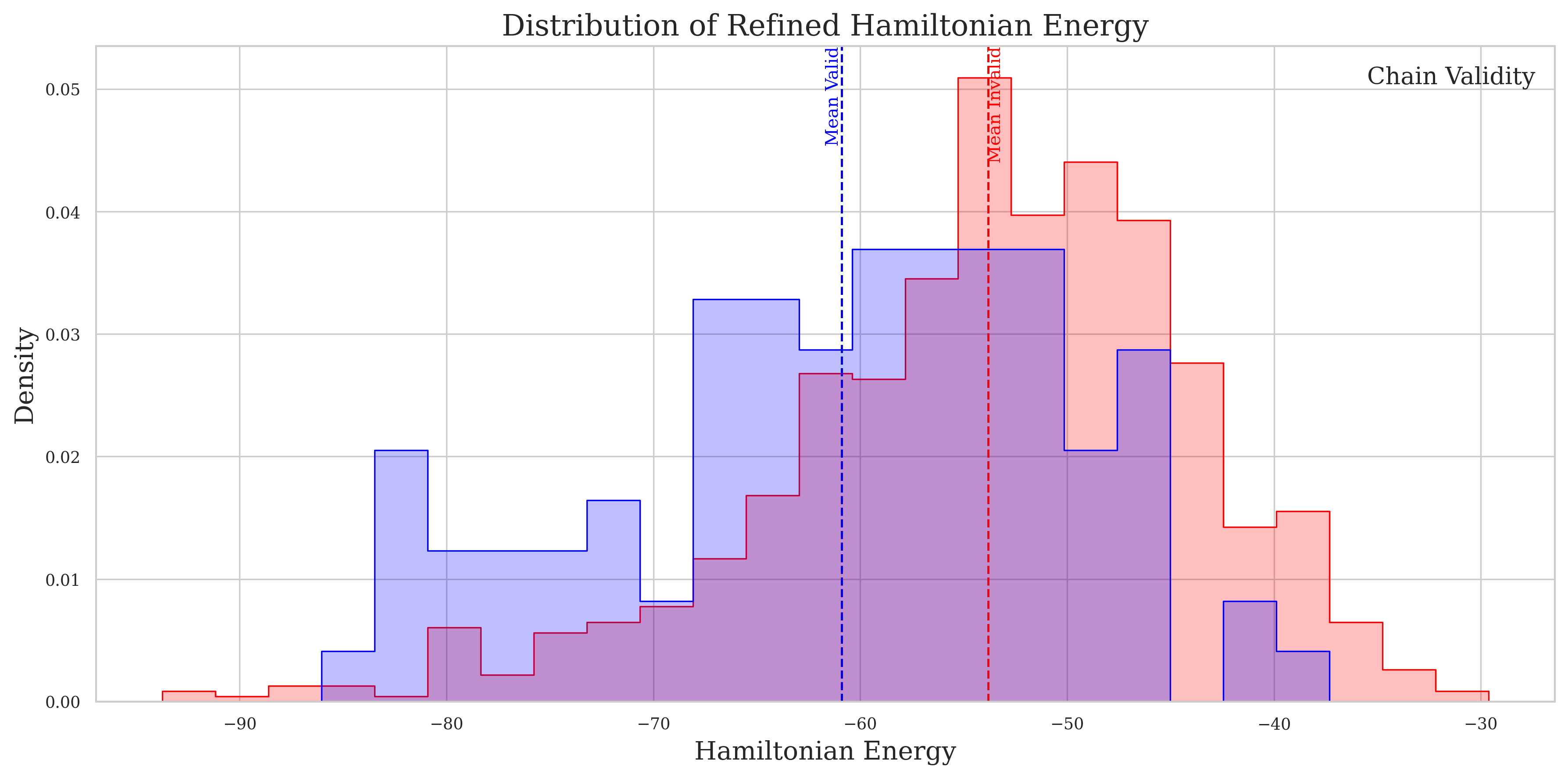}
\caption{Distribution of Hamiltonian energy. Valid chains (blue) and invalid chains (red).}
\label{fig:hamiltonian-energy}
\end{figure}

Figure \ref{fig:hamiltonian-energy} shows the distribution of refined Hamiltonian energy, suggesting that valid chains generally mainly show lower energy states, centering between -60 and -70, while invalid chains show a broader spectrum of energies, reaching higher values. The average energy for valid chains is significantly lower than that of invalid chains. These findings indicate that valid reasoning is characterized by maintaining lower overall energy levels, suggesting an equilibrium between cognitive efficiency and thoroughness in valid reasoning processes. Invalid reasoning tends to involve a broader range of energy levels and rates of change, possibly suggesting less consistent or less efficient cognitive processes.

The Frenet-Serret framework is particularly interesting, as we have already highlighted, because it allows analyzing trajectories geometrically without being dependent on the particular embedding. This is important when working with high-dimensional data that has been reduced to lower dimensions. This framework offers a way to measure the patterns in which reasoning paths twist and curve in abstract space, therefore revealing significant differences between reasoning processes that are valid and invalid.

\begin{figure}[h]
\centering
\includegraphics[width=0.75\textwidth]{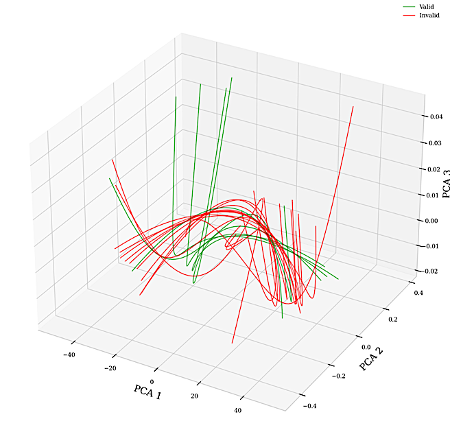}
\caption{Reasoning Trajectories in PCA space using Frenet framework: valid (green) vs invalid chains (red).}
\label{fig:pca-trajectories}
\end{figure}

The PCA space representation of the Frenet-framed reasoning trajectories in Figure \ref{fig:pca-trajectories} shows complex patterns of both valid and invalid chains. Although there appears to be some differences between the two groups based on visual assessment, the statistical tests offer more detailed information. There is a slight overall difference between valid and invalid trajectories across the three PCA dimensions, according to the MANOVA test with a p-value of 0.0830 \cite{olson1979}. According to individual t-tests for each dimension, PCA1 shows a trend toward significance (p = 0.0800), but PCA3 shows clear statistically significant differences (p-value = 0.0496). Although the individual effect sizes of these dimensions are not significant with Cohen's values ranging from 0.0476 to 0.2134 \cite{rosenthal1994}, high accuracy from the logistic regression model's (90.48\%) in classifying chains based on their PCA coordinates suggests the combined discriminatory potential of these dimensions. Specifically, there appear to be different geometric properties between reasoning paths that are valid and invalid based on the difference in trajectory lengths (p = 0.0390). Together, these results show that, although no single geometric feature can
characterize valid and invalid reasoning processes, the combination of Frenet-frame derived properties in lower dimensional space opens a promising way to explore.

\begin{figure}[h]
\centering
\includegraphics[width=0.75\textwidth]{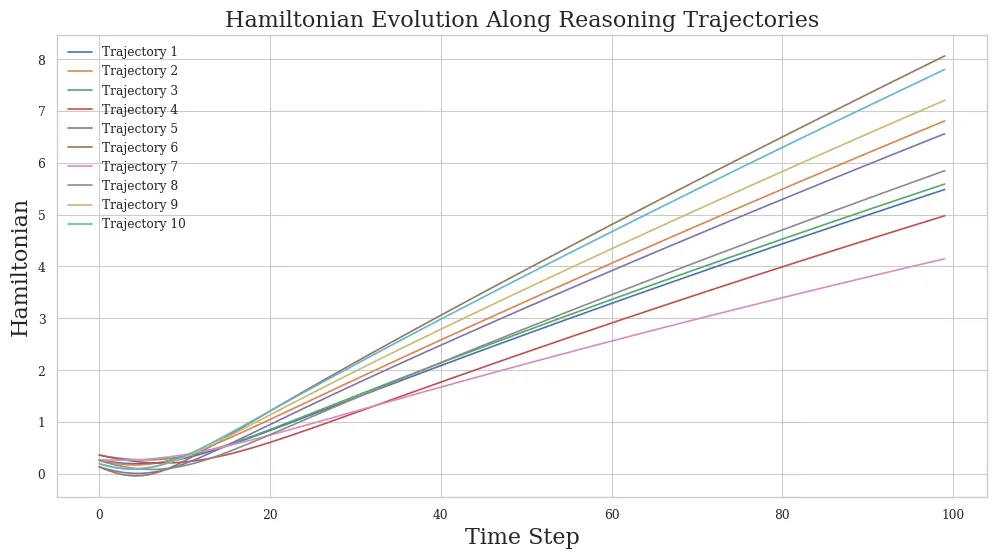}
\caption{Reasoning trajectories evolution trough reasoning process for the first 10 chains within the OBQA dataset.}
\label{fig:trajectories-evolution}
\end{figure}

\begin{table}[h]
\centering
\begin{tabular}{lc}
\hline
\multicolumn{2}{c}{Statistical analysis of reasoning trajectories} \\
\hline
\multicolumn{2}{c}{MANOVA test \quad Multivariate lineal model} \\
Intercept & Value \quad Num DF \quad Den DF \quad F Value \quad Pr > F \\
Wilks' lambda & 0.9954 \quad 3.00 \quad 994.00 \quad 1.5475 \quad 0.2006 \\
Pillai's trace & 0.0046 \quad 3.00 \quad 994.00 \quad 1.5475 \quad 0.2006 \\
Hotelling-Lawley trace & 0.0047 \quad 3.00 \quad 994.00 \quad 1.5475 \quad 0.2006 \\
Roy's greatest root & 0.0047 \quad 3.00 \quad 994.00 \quad 1.5475 \quad 0.2006 \\
\hline
is\_valid & Value \quad Num DF \quad Den DF \quad F Value \quad Pr > F \\
Wilks' lambda & 0.9933 \quad 3.00 \quad 994.00 \quad 2.2313 \quad 0.0830 \\
Pillai's trace & 0.0067 \quad 3.00 \quad 994.00 \quad 2.2313 \quad 0.0830 \\
Hotelling-Lawley trace & 0.0067 \quad 3.00 \quad 994.00 \quad 2.2313 \quad 0.0830 \\
Roy's greatest root & 0.0067 \quad 3.00 \quad 994.00 \quad 2.2313 \quad 0.0830 \\
\hline
t-test for PCA1 & t-statistic \quad 1.7552 \quad p-value \quad 0.0800 \\
t-test for PCA2 & t-statistic \quad 0.4261 \quad p-value \quad 0.6701 \\
t-test for PCA3 & t-statistic \quad 1.9655 \quad p-value \quad 0.0496 \\
Logistic Regression & Accuracy \quad 0.9048 \\
Cohen's d for PCA1 & 0.2006 \\
Cohen's d for PCA2 & 0.0476 \\
Cohen's d for PCA3 & 0.2134 \\
t-test trajectory lenghts & t-statistic \quad -2.0666 \quad p-value \quad 0.0390 \\
\hline
\end{tabular}
\caption{Statistical analysis of reasoning trajectories: MANOVA, PCA, and geometric properties}
\label{tab:trajectories-statistics}
\end{table}

Figure \ref{fig:trajectories-evolution} illustrates the Hamiltonian evolution along reasoning paths. It shows a tangible representation of the temporal variations in the ``energy'' of cognitive reasoning processes, illustrating how different reasoning sequences (both valid and invalid) may exhibit divergent patterns of Hamiltonian evolution. Some trajectories have a more accelerated growth in Hamiltonian value, suggesting more complex reasoning processes.

\begin{figure}[h]
\centering
\includegraphics[width=0.75\textwidth]{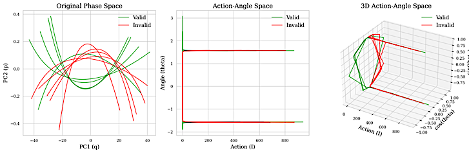}
\caption{Canonical transformations in embedding space of valid vs invalid chains. Original phase space (left), two-dimensional action-phase space (center), and three-dimensional action phase space(right).}
\label{fig:canonical-transformations}
\end{figure}

Figure \ref{fig:canonical-transformations} illustrates the analysis of reasoning paths with the canonical transformations. This approach creates new coordinates $(I, \theta)$, or action-angle variables, by mapping the original phase space coordinates $(q, p)$ to new coordinates. Regarding reasoning paths, $q$ and $p$ can be understood as generalized location and momentum in the cognitive space, representing the reasoning process's present state and rate of change, respectively. The conversion to action-angle variables offers an alternative viewpoint on reasoning dynamics. Similar to energy in physical systems, the action variable $I$ measures the total intensity or complexity of the thought process. It is almost constant along a trajectory, indicating that cognitive resources are conserved or that the degree of engagement is maintained throughout the reasoning activity. Conversely, the phase or direction of reasoning is represented by the angle variable $\theta$, which captures how the approach or focus changes over time.

We can observe in Figure \ref{fig:canonical-transformations} that for both valid and invalid chains, there are complex, nonlinear trajectories in the Original Phase Space. The paths seem to mix, indicating that it is difficult to discern between valid and invalid reasoning in this representation. A strange pattern emerges from the Action-Angle Space: most trajectories are horizontal lines, suggesting that the action $(I)$ is mostly constant while the angle $(\theta)$ fluctuates. This implies that while the ``phase'' or direction of reasoning changes with time, the ``energy'' of reasoning processes tends to be conserved. This trend is further highlighted by the three-dimensional action-angle space, which shows trajectories that mostly rotate around the Action axis. While keeping action values generally constant, the illustration emphasizes the angle variable's periodic nature.

Quantitatively, we find that the mean action of invalid chains is higher (253.4054) than that of valid chains (229.8741). The statistical significance of this difference (t-statistic = -2.5152, p-value = 0.0121) suggests that reasoning processes that are deemed invalid typically have greater ``energy'' or complexity. Nonetheless, there is little statistically significant difference in the mean angle ranges of the valid (3.3961) and non-valid (3.3409) chains (t-statistic = 1.1158, p-value = 0.2648). This implies that the variety of ``phases'' or directions that reasoning processes, whether correct or invalid, traverse in their trajectories is comparable.

These results suggest that, although the reasoning directions covered by valid and non-valid chains may be similar, the reasoning directions covered by non-valid chains are typically more complex or have higher overall energy. This may suggest that while valid reasoning maintains a more efficient energy level while still examining the necessary range of logical processes, non-valid reasoning often involves more complex or convoluted paths, potentially increasing reasoning process complexity.

The Hamiltonian evolution and the analysis of canonical transformations provide different yet complementary insights into the dynamics of reasoning. Figure \ref{fig:trajectories-evolution} illustrates the progressive increase of reasoning processes over time, whereas the action-angle representation shows, through the canonical transformations, a more detailed perspective. The consistent action values observed during the transformation indicate that the overall complexity or ``energy'' of reasoning tends to remain stable, in contrast to the progressively rising Hamiltonian values. Canonical transformations are very good at showing conserved quantities that aren't obvious in the original representation because of this apparent inconsistency. The angle variable, which indicates the phase or direction of reasoning, shows comparable ranges for both valid and invalid chains. This suggests that the effectiveness of energy utilization, rather than the extent of cognitive exploration, drives the differentiation between them. This observation, which is not evident from the Hamiltonian evolution alone, illustrates how the action-angle framework enhances comprehension of the fundamental dynamics, uncovering complex differences in reasoning processes that could be essential for differentiating between valid and invalid chains in AI systems.

\subsection{Conservation laws in reasoning dynamics}

\begin{figure}[h]
\centering
\includegraphics[width=0.75\textwidth]{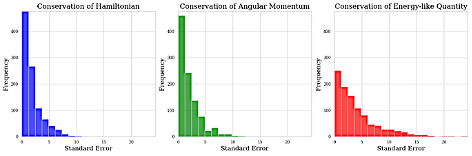}
\caption{Conservation Laws in Reasoning Trajectories: Hamiltonian (left), Angular Momentum (middle), and Energy-like Quantity (right).}
\label{fig:conservation-laws}
\end{figure}

Hamiltonian conservation (blue color in Figure \ref{fig:conservation-laws}) shows a pronounced skew towards smaller standard error values. A significant fraction of trajectories shows few alterations (high conservation). As the standard error increases, the frequency decreases rapidly. This suggests that numerous reasoning paths consistently preserve the Hamiltonian. The plot illustrating the conservation of angular momentum (green) is analogous to the Hamiltonian, however exhibiting a little broader distribution. The plot consistently shows a significant trend towards lower standard error levels. The decrease in frequency as the standard error rises is less pronounced than that of the Hamiltonian. This suggests effective conservation of angular momentum, although maybe less strict than the Hamiltonian. Finally, in the plot showing a quantity analogous to energy conservation (Red and Blue colors in Figure \ref{fig:conservation-laws}), the distribution is far broader than the other two. Although a peak persists at lower standard error levels, it is less prominent. The distribution's tail stretches wider, indicating an increased number of cases with bigger variations. This suggests that the energy-like quantity conservation is less consistent than that of the Hamiltonian and angular momentum.

The plots confirm a different conservation structure. The Hamiltonian is the most rigorously conserved, closely followed by angular momentum, whereas the energy-like quantity shows the lowest level of conservation. Hamiltonian conservation explains that some reasoning paths maintain a stable overall ``energy'' or complexity over their progression. The broader distribution of the energy-like amount may suggest that this component of reasoning has greater flexibility or variability over diverse paths. These distributions' unique patterns may be used in the classification of different thinking styles. Trajectories with minimal standard errors across all parameters indicate very stable or consistent reasoning paths. All quantities show different levels of conservation, supporting the idea that there are fundamental principles or limitations governing the dynamics of reasoning processes. These differing levels of conservation offer empirical evidence for enhancing our theoretical framework. Strong Hamiltonian conservation parallels physical systems, validating this framework as an explanatory model for reasoning processes.

Different conservation levels across Hamiltonian, angular momentum, and energy-like quantities suggest underlying symmetries in LLMs reasoning processes. The strongly conserved Hamiltonian indicates a fundamental invariance in reasoning structures, similar to time-translation symmetry in physics. Less strictly conserved quantities point to approximate symmetries in cognitive processes. These findings validate our Hamiltonian approach while opening new paths to explore symmetries in LLMs. Applying Noether's theorem to reasoning processes provides a framework for understanding fundamental principles of human and artificial intelligence, potentially advancing more robust reasoning models.

\subsection{Geometric Analysis of Reasoning Trajectories}

Differential geometry applied to reasoning trajectories offers a robust framework for analyzing reasoning structures and properties, allowing us to characterize cognitive paths through high-dimensional space.

\begin{figure}[h]
\centering
\includegraphics[width=0.75\textwidth]{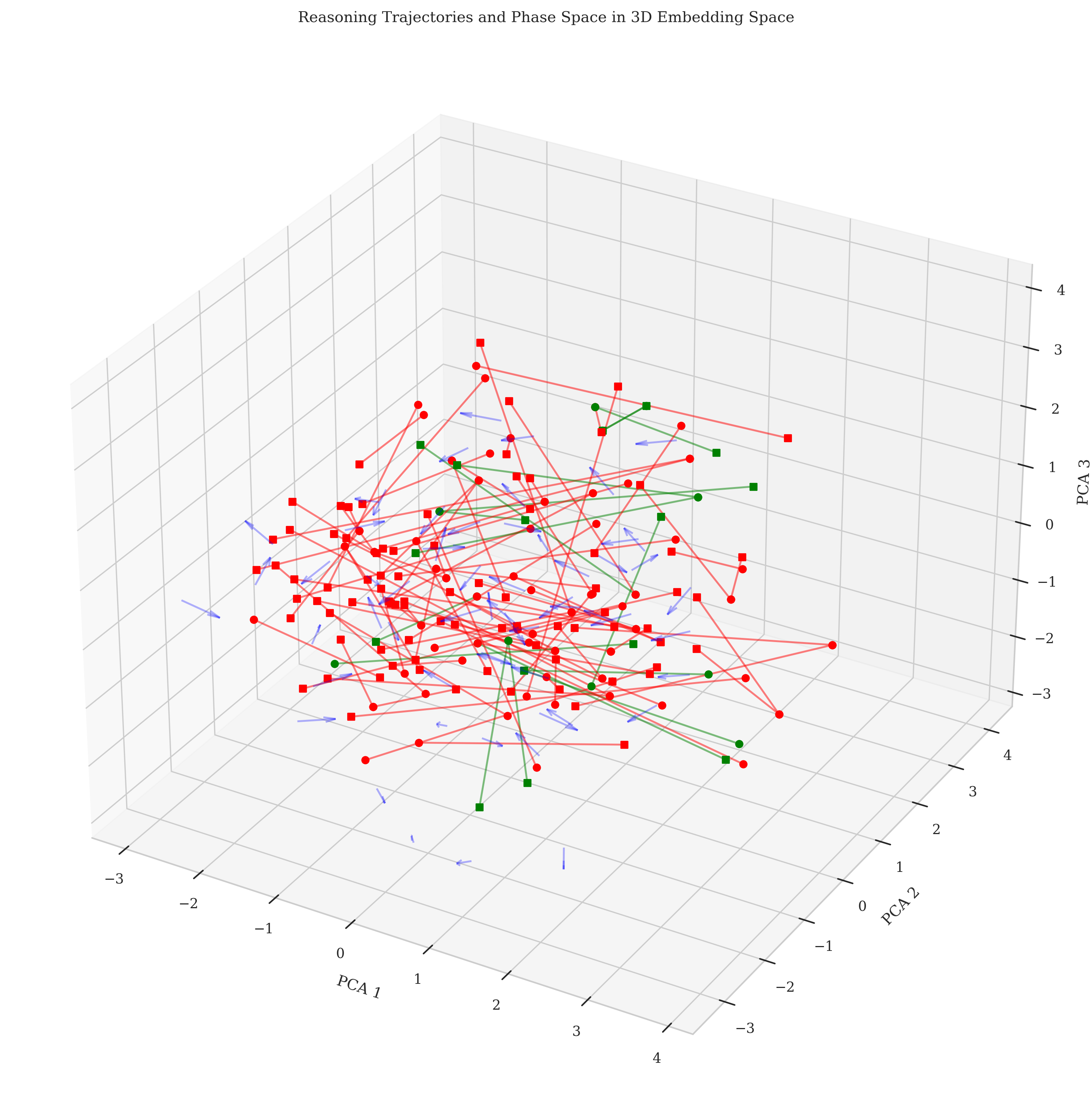}
\caption{Reasoning trajectories within OBQA dataset mapped in a three dimensional space. Valid chains (green), invalid chains (red).}
\label{fig:3d-trajectories}
\end{figure}

Figure \ref{fig:3d-trajectories} illustrates PCA-reduced reasoning paths from the OBQA dataset, revealing complex cognitive structures. The densely populated space shows invalid chains (red) dominating with chaotic, scattered patterns, while valid chains (green) appear less frequently but follow more focused, directed trajectories. Blue arrows indicate general reasoning flow. Trajectory clusters in the core region suggest common reasoning patterns, while outlying paths represent unusual cognitive processes. Where valid and invalid paths converge, subtle distinctions emerge. This visualization demonstrates reasoning complexity and the challenge of distinguishing valid from invalid reasoning through trajectory patterns alone.

\begin{figure}[h]
\centering
\includegraphics[width=0.75\textwidth]{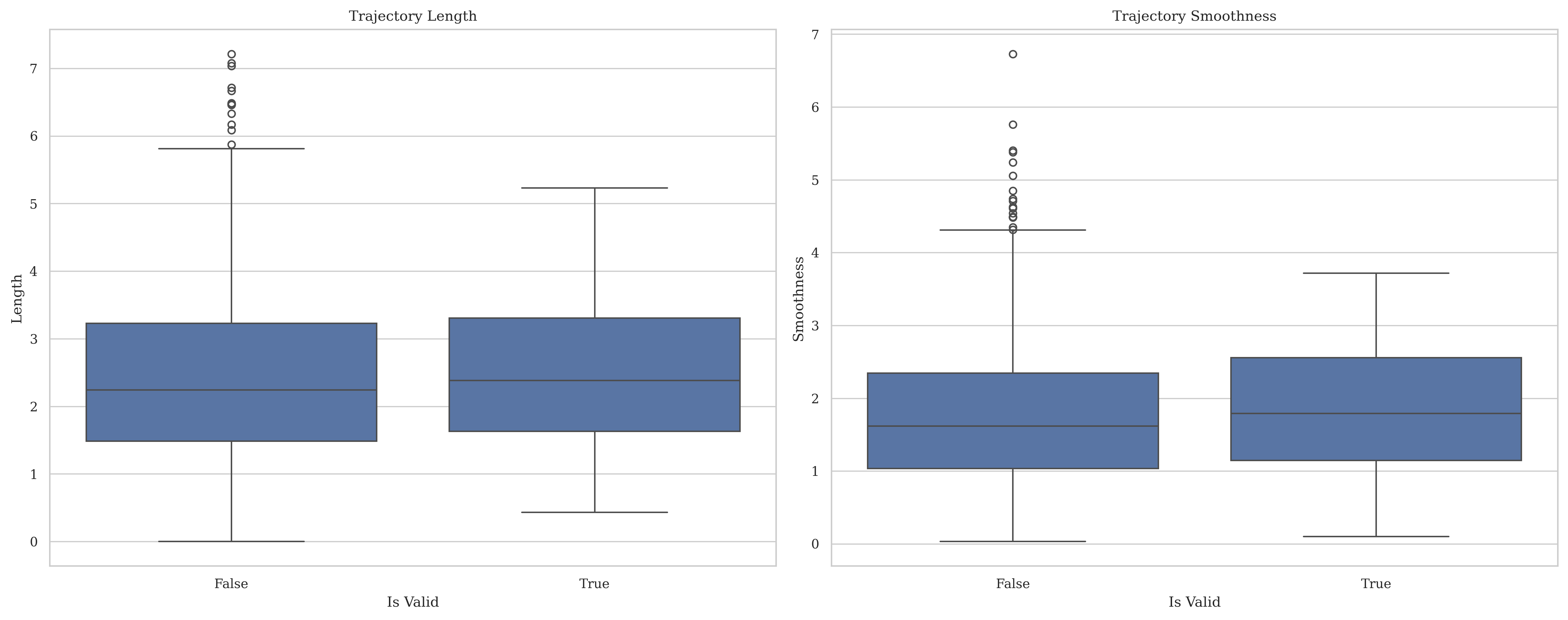}
\caption{Trajectory length (left) and smoothness (right) within OBQA dataset.}
\label{fig:trajectory-properties}
\end{figure}

Figure \ref{fig:trajectory-properties} quantifies two key features of reasoning chains: length and smoothness. The box plots show that valid and invalid trajectories have similar length distributions (comparable medians and inter-quartile ranges), confirming our observation from the three-dimensional visualization where both types appeared mixed. However, smoothness significantly differs - valid trajectories demonstrate greater smoothness with higher median values and narrower distributions, corresponding to the more focused green trajectories in the three-dimensional plot. Outliers appear in both plots but are more prevalent among invalid chains, reflecting their erratic patterns. These metrics confirm that while path length doesn't reliably indicate validity, the smoothness of cognitive movement through conceptual space better distinguishes valid reasoning processes.

\begin{figure}[h]
\centering 
\includegraphics[width=0.75\textwidth]{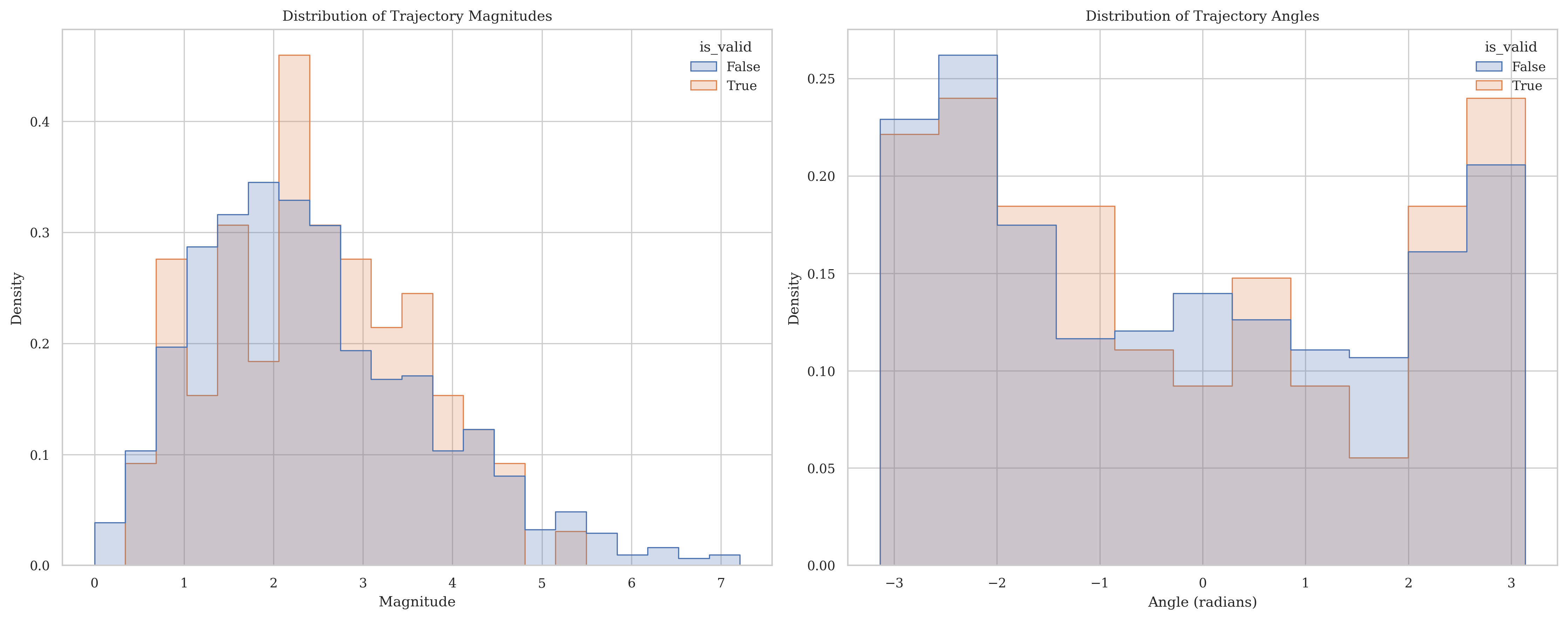}
\caption{Magnitude (left) and angle (right) distributions of reasoning trajectories within OBQA dataset. Valid chains (orange), invalid chains (blue).}
\label{fig:magnitude-angle}
\end{figure}

Figure \ref{fig:magnitude-angle} shows trajectory magnitude and angle distributions for valid and invalid reasoning chains. Valid chains (orange) have slightly higher magnitudes peaking between 2-3, consistent with their more directed trajectories in three-dimensional space. Invalid chains (blue) show wider magnitude dispersion, matching their scattered three-dimensional representation. For angles, valid chains cluster near $\pi$ radians, indicating more uniform directional changes, while invalid chains distribute more evenly across angles, suggesting irregular patterns. These distributions confirm our box plot findings: trajectory smoothness differs slightly between valid and invalid chains, while trajectory length alone cannot reliably identify valid chains.

\begin{figure}[h]
\centering 
\includegraphics[width=0.75\textwidth]{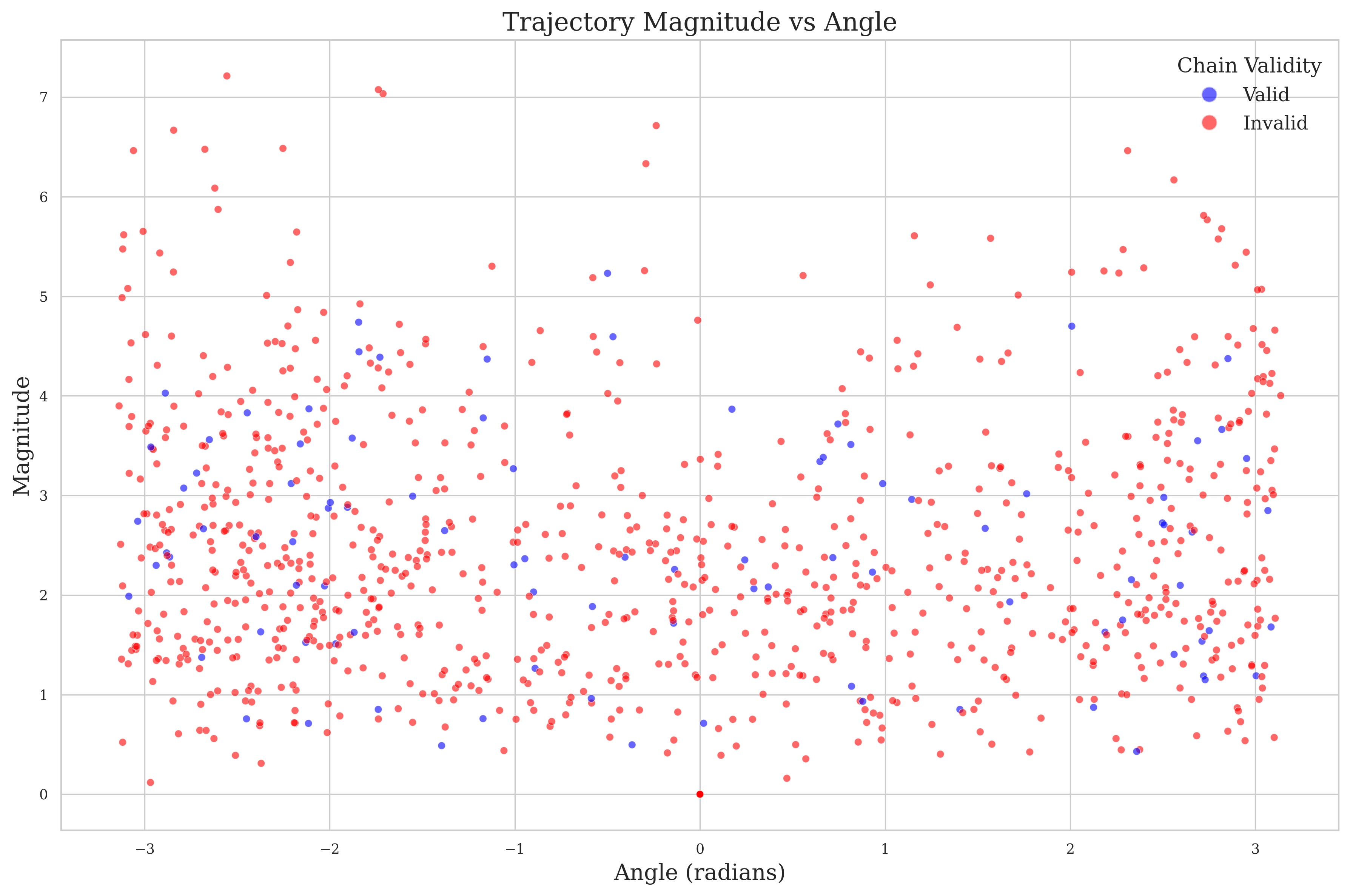}
\caption{Trajectory magnitude vs angle of reasoning trajectories of valid/invalid chains within OBQA dataset.}
\label{fig:magnitude-vs-angle}
\end{figure}

These visualizations reveal characteristic differences in reasoning patterns: valid chains exhibit more uniform magnitudes and angles, creating smoother, more directed trajectories through conceptual space, while invalid chains display greater variability, resulting in scattered, less predictable paths.

\begin{table}[h]
\centering
\begin{tabular}{lc}
\hline
\multicolumn{2}{c}{Statistical tests for different features} \\
\hline
Average Energy Conservation Score & 0.019 \\
Correlation Energy Conservation vs. Validity & -0.189 \\
\hline
Correlation between Angle and Magnitude (overall) & -0.029 \\
Correlation for Valid Chains & -0.129 \\
Correlation for Invalid Chains & -0.020 \\
\hline
\multicolumn{2}{c}{T-test for difference in Trajectory Angle} \\
t-statistic & 0.1439 \\
p-value & 0.8856 \\
\hline
\multicolumn{2}{c}{T-test for difference in Trajectory Magnitude} \\
t-statistic & 0.2467 \\
p-value & 0.8052 \\
\hline
\end{tabular}
\caption{Geometric properties statistics of reasoning chains}
\label{tab:geometric-properties}
\end{table}

\begin{figure}[h]
\centering
\includegraphics[width=0.75\textwidth]{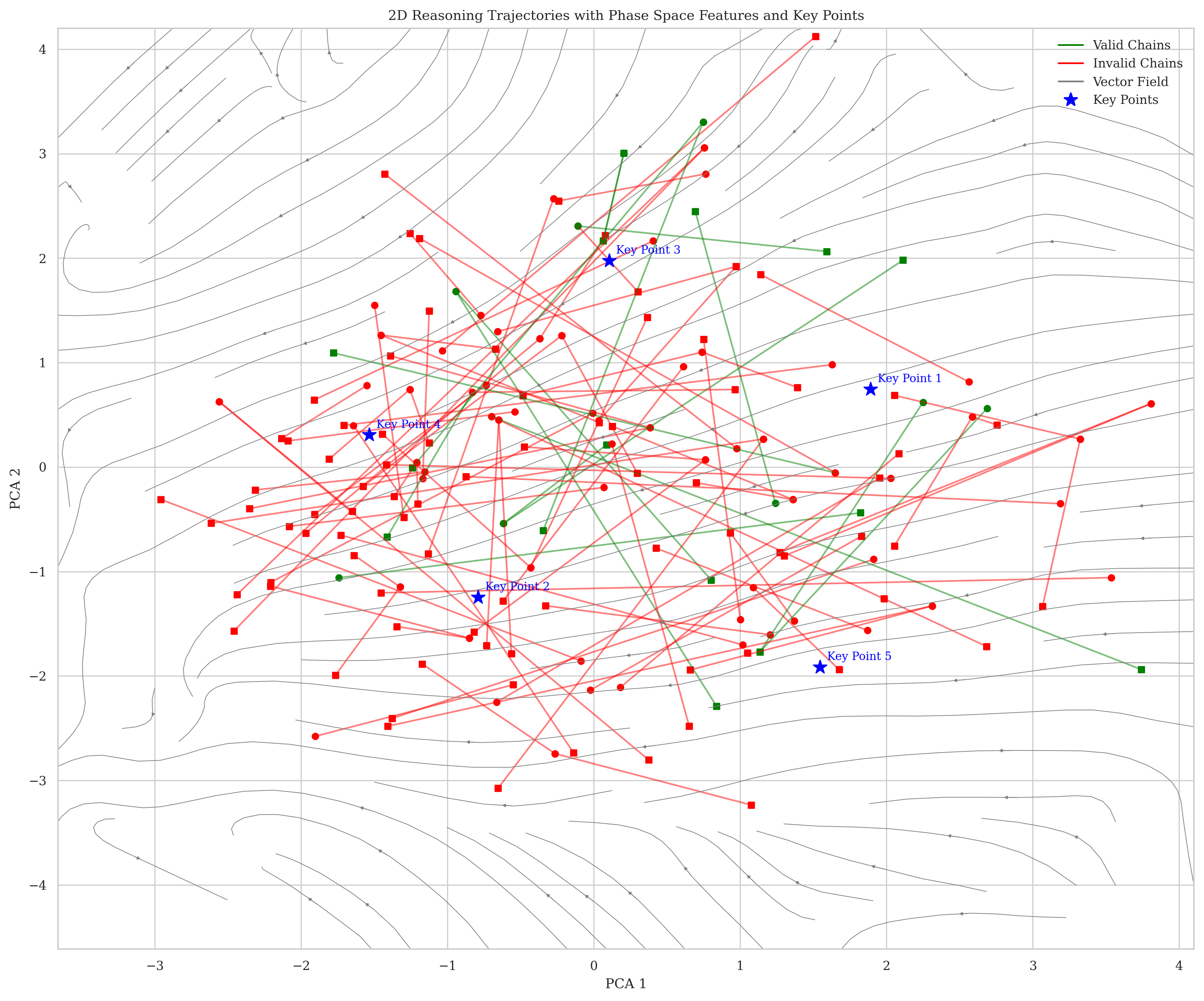}
\caption{Two-dimensional reasoning trajectories within OBQA dataset.}
\label{fig:2d-trajectories}
\end{figure}

Figure \ref{fig:2d-trajectories} represents a two-dimensional projection of reasoning paths, complementing our previous three-dimensional analysis while revealing temporal dynamics in the OBQA dataset. This visualization combines phase space features with critical points for enhanced analytical clarity. Valid (green) and invalid (red) reasoning chains appear as trajectories in the space defined by the first two principal components. Gray arrows form a vector field showing the general flow of reasoning, comparable to energy contours in Figure \ref{fig:phase-plots}, while blue stars mark stable states or key concepts representing pivotal transformations in reasoning.
The trajectories exhibit both focused problem solving and multi-concept reasoning patterns. Valid chains often follow more constrained and directed paths, suggesting deep engagement with specific topics. Some invalid chains cover broader conceptual areas, potentially indicating multi-concept reasoning with less coherence.
Vector field convergence and divergence regions likely correspond to equilibrium points where significant conceptual shifts occur. While this two-dimensional representation clarifies the overall reasoning structure at the expense of some detail, it effectively bridges our abstract three-dimensional visualization with theoretical phase space concepts, demonstrating how OBQA dataset reasoning sequences align with Hamiltonian systems in both focused and multi-concept reasoning modes.

\subsection{Statistical mechanics and computational analysis of reasoning trajectories}

\begin{figure}[h]
\centering
\includegraphics[width=0.75\textwidth]{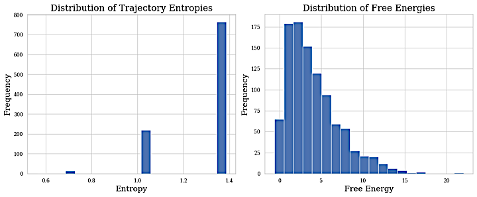}
\caption{Analysis of entropy (left) and free energy (right) in reasoning trajectories within OBQA dataset. Mean trajectory Entropy=1.3004 ; Mean free energy=4.0237}
\label{fig:entropy-free-energy}
\end{figure}

We analyzed trajectory entropy and free energy distributions (Figure \ref{fig:entropy-free-energy}), finding a strong bias toward higher entropy values (primarily around 1.4, with a minor peak at 1.0). This indicates widespread instability or unpredictability in reasoning paths. The more dispersed free energy distribution peaks between 2-3 with an extended tail toward higher values, revealing varying degrees of stability across reasoning processes, with some pathways being more energetically favorable than others.

\begin{figure}[h]
\centering
\includegraphics[width=0.75\textwidth]{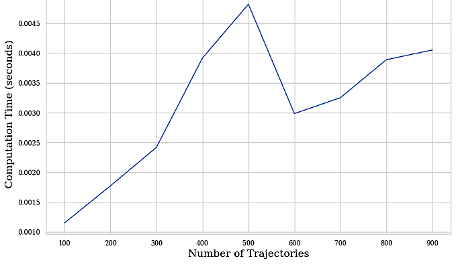}
\caption{Calculation of computational complexity. Estimated complexity: $O(n^{0.43})$}
\label{fig:computational-complexity}
\end{figure}

Figure \ref{fig:computational-complexity} shows a non-linear increase in computation time as the number of trajectories increases. The estimated complexity of $O(n^{0.43})$ suggests that our algorithm scales sub-linearly with the number of trajectories. This is quite efficient and indicates high scalability for larger datasets.

\begin{figure}[h]
\centering
\includegraphics[width=0.75\textwidth]{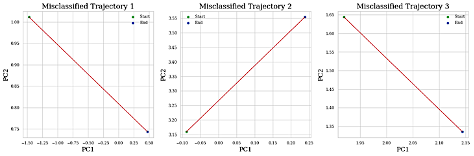}
\caption{Classification of reasoning trajectories within OBQA dataset.}
\label{fig:classification}
\end{figure}

Figure \ref{fig:classification} presents three examples of misclassified reasoning chains, ranging from nearly linear trajectories to complex paths in PC1-PC2 space. This visualization shows potential classification error, including similar endpoints or unusual path geometries. Our model effectively identifies invalid chains (0.91 precision, 0.83 recall) but performs poorly on valid chains (0.18 precision, 0.33 recall). Overall accuracy of 0.78 indicates significant improvement opportunities, particularly for valid chain identification.

\begin{table}[h]
\centering
\begin{tabular}{cc}
\hline
\multicolumn{2}{c}{Confusion matrix} \\
148 & 31 \\
14 & 7 \\
\hline
\end{tabular}
\caption{Confusion matrix for classification of valid and invalid chains}
\label{tab:confusion-matrix}
\end{table}

\begin{table}[h]
\centering
\begin{tabular}{lccc}
\hline
\multicolumn{4}{c}{Classification Report} \\
& precision & recall & f1-score \\
\hline
False & 0.91 & 0.83 & 0.87 \\
True & 0.18 & 0.33 & 0.24 \\
\hline
accuracy & & & 0.78 \\
macro avg & 0.55 & 0.58 & 0.55 \\
weighted avg & 0.84 & 0.78 & 0.80 \\
\hline
\end{tabular}
\caption{Report for classification of valid and invalid chains}
\label{tab:classification-report}
\end{table}

\section{Discussion}

Applying Hamiltonian mechanics and differential geometry to reasoning trajectories reveals key insights for understanding and improving AI reasoning. This framework offers a novel perspective on cognitive dynamics in embedding spaces with important theoretical and practical implications.

\subsection{Interpretation of key findings}

Valid reasoning chains show lower Hamiltonian energy levels than invalid chains, suggesting effective reasoning achieves a more efficient balance between the "kinetic" energy of state transitions and the "potential" energy of semantic relevance. Invalid chains show wider energy ranges with higher values, suggesting ineffective reasoning may involve less stable or more energy-intensive cognitive transitions.

Trajectory geometry analysis shows valid reasoning follows smoother paths with lower curvature, indicating direct and focused conceptual progression. Invalid chains show higher curvature and torsion, suggesting convoluted or incoherent reasoning. We identified quantities analogous to physical conservation laws in reasoning trajectories, with stronger conservation in valid chains pointing to fundamental invariances in model effective reasoning processes, similar to physical systems.

Converting reasoning trajectories to action-angle variables demonstrates that while reasoning tends to conserve "action" (energy-like quantity), the "angle" (conceptual direction) varies freely. This supports the intuition that effective reasoning maintains consistent engagement complexity while exploring diverse cognitive directions.

\subsection{Implications for LLMs reasoning}

This Hamiltonian approach provides novel quantification and visualization of reasoning processes, potentially improving AI system explainability by representing reasoning as physical-like trajectories. While conceptually elegant, we must acknowledge several limitations to this framework's practical utility.

The mapping between physical systems and reasoning processes remains largely metaphorical rather than established. There's limited empirical evidence that optimizing for lower-energy or smoother trajectories would meaningfully improve reasoning performance beyond what existing methods achieve. The abstract nature of embedding spaces also presents challenges in translating theoretical insights into actionable algorithm improvements.

Nevertheless, this framework may offer valuable diagnostic capabilities. Unusual trajectory patterns or anomalous energy distributions could identify problematic reasoning, potentially helping detect biases or logical inconsistencies. The geometric representation of reasoning might also provide a shared conceptual framework for comparing human and LLM reasoning processes, though claims of direct parallels should be approached with caution until more thoroughly validated.

\subsection{Limitations and challenges}

This study's scope is limited by its reliance on the OBQA dataset alone, requiring validation across diverse reasoning tasks and domains. PCA visualization, while necessary for interpretation, likely obscures important high-dimensional nuances; alternative dimension reduction techniques like t-SNE should be explored. We must guard against drawing unwarranted conclusions from the physical-cognitive analogy, ensuring observed patterns represent genuine reasoning dynamics rather than mathematical artifacts of our formalism.

\subsection{Future directions}

\subsection{Future directions}

We propose several key research directions: (1) extending this framework to diverse reasoning tasks and larger datasets to assess domain-general applicability and identify task-specific patterns; (2) developing real-time trajectory analysis methods for dynamic intervention in LLM reasoning; (3) incorporating energy conservation principles and geometric constraints into neural network architectures; (4) exploring quantum mechanical analogies for uncertainty management and concept superposition; (5) collaborating with cognitive scientists to investigate parallels between AI and human reasoning; and (6) creating accessible tools for implementing this framework.

This integration of Hamiltonian mechanics and differential geometry into AI reasoning represents a promising approach for quantifying, visualizing, and potentially optimizing reasoning processes. Further refinement could enhance LLM performance while providing deeper insights into the fundamental nature of reasoning and cognition of these models.


\begin{thebibliography}{99}

\bibitem{amari2016} Amari, S. (2016). Information geometry and its applications (Vol. 194). Springer.

\bibitem{andersen1972} Andersen, P. W. (1972). More is different. Science, 177(4047), 393--396.

\bibitem{arnold2013} Arnold, D., \& Igorevich, V. (2013). Mathematical methods of classical mechanics (Vol. 60). Springer Science \& Business Media.

\bibitem{barabasi1999} Barabási, A. L., \& Albert, R. (1999). Emergence of Scaling in Random Networks. Science, 286(5439), 509--512.

\bibitem{barto2013} Barto, A. G. (2013). Intrinsic Motivation and Reinforcement Learning. Intrinsically Motivated Learning in Natural and Artificial Systems, 9783642323751, 17--47.

\bibitem{beaty2014} Beaty, R. E., Benedek, M., Wilkins, R. W., Jauk, E., Fink, A., Silvia, P. J., Hodges, D. A., Koschutnig, K., \& Neubauer, A. C. (2014). Creativity and the default network: A functional connectivity analysis of the creative brain at rest. Neuropsychologia, 64, 92--98.

\bibitem{beaty2018} Beaty, R. E., Kenett, Y. N., Christensen, A. P., Rosenberg, M. D., Benedek, M., Chen, Q., Fink, A., Qiu, J., Kwapil, T. R., \& Kane, M. J. (2018). Robust prediction of individual creative ability from brain functional connectivity. Proceedings of the National Academy of Sciences, 115(5), 1087--1092.

\bibitem{bender2021} Bender, E. M., Gebru, T., McMillan-Major, A., \& Shmitchell, S. (2021). On the dangers of stochastic parrots: Can language models be too big? FAccT 2021 - Proceedings of the 2021 ACM Conference on Fairness, Accountability, and Transparency, 610--623.

\bibitem{bengio2013} Bengio, Y., Courville, A., \& Vincent, P. (2013). Representation learning: A review and new perspectives. IEEE Transactions on Pattern Analysis and Machine Intelligence, 35(8), 1798--1828.

\bibitem{bengio2021} Bengio, Y., Lecun, Y., \& Hinton, G. (2021). Deep learning for AI. Communications of the ACM, 64(7), 58--65.

\bibitem{bertsimas1993} Bertsimas, D., \& Tsitsiklis, J. (1993). Simulated annealing. Statistical Science, 8(1), 10--15.

\bibitem{bialek2012} Bialek, W. (2012). Biophysics: searching for principles. Princeton University Press.

\bibitem{carleo2019} Carleo, G., Cirac, I., Cranmer, K., Daudet, L., Schuld, M., Tishby, N., Vogt-Maranto, L., \& Zdeborová, L. (2019). Machine learning and the physical sciences. Reviews of Modern Physics, 91(4), 045002.

\bibitem{chen2019} Chen, J., \& Durrett, G. (2019). Understanding dataset design choices for multi-hop reasoning. ArXiv Preprint ArXiv:1904.12106.

\bibitem{de2011} De León, M., \& Rodrigues, P. R. (2011). Generalized Classical Mechanics and Field Theory: a geometrical approach of Lagrangian and Hamiltonian formalisms involving higher order derivatives. Elsevier.

\bibitem{devlin2018} Devlin, J., Lee, K., Chang, M., \& Toutanova, K. (2018). Bert: Pre-training of deep bidirectional transformers for language understanding. ArXiv Preprint ArXiv:1810.04805.

\bibitem{do2016} Do Carmo, M. P. (2016). Differential geometry of curves and surfaces: revised and updated second edition. Courier Dover Publications.

\bibitem{dong2023} Dong, J., Zhang, Q., Huang, X., Duan, K., Tan, Q., \& Jiang, Z. (2023). Hierarchy-Aware Multi-Hop Question Answering over Knowledge Graphs. ACM Web Conference 2023 - Proceedings of the World Wide Web Conference, WWW 2023, 2519--2527.

\bibitem{dua2019} Dua, D., Wang, Y., Dasigi, P., Stanovsky, G., Singh, S., \& Gardner, M. (2019). DROP: A reading comprehension benchmark requiring discrete reasoning over paragraphs. ArXiv Preprint ArXiv:1903.00161.

\bibitem{duistermaat2012} Duistermaat, J. J., \& Kolk, J. A. C. (2012). Lie groups. Springer Science \& Business Media.

\bibitem{easton1993} Easton, R. W. (1993). Introduction to Hamiltonian dynamical systems and the N-body problem (KR Meyer and GR Hall). SIAM Review, 35(4), 659.

\bibitem{elsage2022} Elsage, N., Olsson, C., Amodei, D., Olah, C., \& Kaplan, J. (2022). Toy Models of Superposition. Transformer Circuits Thread.

\bibitem{feng2020} Feng, Y., Chen, X., Lin, B. Y., Wang, P., Yan, J., \& Ren, X. (2020). Scalable Multi-Hop Relational Reasoning for Knowledge-Aware Question Answering. EMNLP 2020 - 2020 Conference on Empirical Methods in Natural Language Processing, Proceedings of the Conference, 1295--1309.

\bibitem{feynman1967} Feynman, R. (1967). The character of physical law (1965). Cox and Wyman Ltd., London.

\bibitem{friston2010} Friston, K. (2010). The free-energy principle: a unified brain theory? Nature Reviews Neuroscience, 11(2), 127--138.

\bibitem{fujimoto2001} Fujimoto, K., \& Sugie, T. (2001). Canonical transformation and stabilization of generalized Hamiltonian systems. Systems \& Control Letters, 42(3), 217--227.

\bibitem{garbuio2021} Garbuio, M., \& Lin, N. (2021). Innovative idea generation in problem finding: Abductive reasoning, cognitive impediments, and the promise of artificial intelligence. Journal of Product Innovation Management, 38(6), 701--725.

\bibitem{geirhos2020} Geirhos, R., Jacobsen, J. H., Michaelis, C., Zemel, R., Brendel, W., Bethge, M., \& Wichmann, F. A. (2020). Shortcut learning in deep neural networks. Nature Machine Intelligence 2020 2:11, 2(11), 665--673.

\bibitem{glattfelder2019} Glattfelder, J. B. (2019). The Semantics of Symmetry, Invariance, and Structure. Frontiers Collection, Part F1071, 65--92.

\bibitem{goldman1967} Goldman. (1967). A causal theory of knowing. Journal of Philosophy, 64(12), 357--372.

\bibitem{goldman1984} Goldman, W. M. (1984). The symplectic nature of fundamental groups of surfaces. Advances in Mathematics, 54(2), 200--225.

\bibitem{goldstein2002} Goldstein, P., \& Poole, C. (2002). Classical Mechanics. Addison Wesley.

\bibitem{hairer2006} Hairer, E., Hochbruck, M., Iserles, A., \& Lubich, C. (2006). Geometric Numerical Integration. Oberwolfach Reports, 3(1), 805--882.

\bibitem{helie2010} Hélie, S., \& Sun, R. (2010). Incubation, insight, and creative problem solving: a unified theory and a connectionist model. Psychological Review, 117(3), 994.

\bibitem{hirsch2013} Hirsch, M. W., Smale, S., \& Devaney, R. L. (2013). Differential equations, dynamical systems, and an introduction to chaos. Academic press.

\bibitem{ho2020} Ho, X., Nguyen, A. K. D., Sugawara, S., \& Aizawa, A. (2020). Constructing A Multi-hop QA Dataset for Comprehensive Evaluation of Reasoning Steps. COLING 2020 - 28th International Conference on Computational Linguistics, Proceedings of the Conference, 6609--6625.

\bibitem{jarvela2023} Järvelä, S., Nguyen, A., \& Hadwin, A. (2023). Human and artificial intelligence collaboration for socially shared regulation in learning. British Journal of Educational Technology, 54(5), 1057--1076.

\bibitem{jhamtani2020a} Jhamtani, H., \& Clark, P. (2020). Learning to Explain: Datasets and Models for Identifying Valid Reasoning Chains in Multihop Question-Answering. EMNLP 2020 - 2020 Conference on Empirical Methods in Natural Language Processing, Proceedings of the Conference, 137--150.

\bibitem{jhamtani2020b} Jhamtani, H., \& Clark, P. (2020). Learning to explain: Datasets and models for identifying valid reasoning chains in multihop question-answering. ArXiv Preprint ArXiv:2010.03274.

\bibitem{jolliffe2016} Jolliffe, I. T., \& Cadima, J. (2016). Principal component analysis: a review and recent developments. Philosophical Transactions of the Royal Society A: Mathematical, Physical and Engineering Sciences, 374(2065), 20150202.

\bibitem{kaelbling1996} Kaelbling, L. P., Littman, M. L., \& Moore, A. W. (1996). Reinforcement Learning: A Survey. Journal of Artificial Intelligence Research, 4, 237--285.

\bibitem{karasev2012} Karasev, M. V., \& Maslov, V. P. (2012). Nonlinear Poisson brackets: geometry and quantization (Vol. 119). American Mathematical Soc.

\bibitem{khot2020} Khot, T., Clark, P., Guerquin, M., Jansen, P., \& Sabharwal, A. (2020). QASC: A Dataset for Question Answering via Sentence Composition. Proceedings of the AAAI Conference on Artificial Intelligence, 34(05), 8082--8090.

\bibitem{kosmann2011} Kosmann-Schwarzbach, Y., Schwarzbach, B. E., \& KosmannSchwarzbach, Y. (2011). The noether theorems. Springer.

\bibitem{kounios2014a} Kounios, J., \& Beeman, M. (2014). The cognitive neuroscience of insight. Annual Review of Psychology, 65(1), 71--93.

\bibitem{kounios2014b} Kounios, J., \& Beeman, M. (2014). The cognitive neuroscience of insight. Annual Review of Psychology, 65(1), 71--93.

\bibitem{lipton2018} Lipton, Z. C. (2018). The mythos of model interpretability: In machine learning, the concept of interpretability is both important and slippery. Queue, 16(3), 31--57.

\bibitem{maruthi2022} Maruthi, S., Dodda, S. B., Yellu, R. R., Thuniki, P., Reddy, S., \& Reddy, B. (2022). Temporal Reasoning in AI Systems: Studying temporal reasoning techniques and their applications in AI systems for modeling dynamic environments. Journal of AI-Assisted Scientific Discovery, 2(2), 22--28.

\bibitem{mehrabi2021} Mehrabi, N., Morstatter, F., Saxena, N., Lerman, K., \& Galstyan, A. (2021). A Survey on Bias and Fairness in Machine Learning. ACM Computing Surveys (CSUR), 54(6).

\bibitem{mehta2019} Mehta, P., Bukov, M., Wang, C. H., Day, A. G. R., Richardson, C., Fisher, C. K., \& Schwab, D. J. (2019). A high-bias, low-variance introduction to Machine Learning for physicists. Physics Reports, 810, 1--124.

\bibitem{mihaylov2018} Mihaylov, T., Clark, P., Khot, T., \& Sabharwal, A. (2018). Can a suit of armor conduct electricity? a new dataset for open book question answering. ArXiv Preprint ArXiv:1809.02789.

\bibitem{mikolov2013a} Mikolov, T., Sutskever, I., Chen, K., Corrado, G. S., \& Dean, J. (2013). Distributed Representations of Words and Phrases and their Compositionality. Advances in Neural Information Processing Systems, 26.

\bibitem{mikolov2013b} Mikolov, T., Sutskever, I., Chen, K., Corrado, G. S., \& Dean, J. (2013). Distributed Representations of Words and Phrases and their Compositionality. Advances in Neural Information Processing Systems, 26.

\bibitem{olson1979} Olson, C. L. (1979). Practical considerations in choosing a MANOVA test statistic: a rejoinder to Stevens.

\bibitem{oneill2006} O'neill, B. (2006). Elementary differential geometry. Elsevier.

\bibitem{pennington2014} Pennington, J., Socher, R., \& Manning, C. D. (2014). Glove: Global vectors for word representation. Proceedings of the 2014 Conference on Empirical Methods in Natural Language Processing (EMNLP), 1532--1543.

\bibitem{penrose2006} Penrose, R. (2006). The road to reality. Random house.

\bibitem{popper1959} Popper, K. (1959). The logic of scientific discovery. Routledge.

\bibitem{prugovecki1979} Prugovečki, E. (1979). Stochastic phase spaces and master Liouville spaces in statistical mechanics. Foundations of Physics, 9(7--8), 575--587.

\bibitem{quine1948} Quine, W. V. O. (1948). On what there is. Catholic University of America, Philosophy Education Society Washington, DC.

\bibitem{rahutomo2012} Rahutomo, F., Kitasuka, T., \& Aritsugi, M. (2012). Semantic cosine similarity. The 7th International Student Conference on Advanced Science and Technology ICAST, 4(1), 1.

\bibitem{reimers2019} Reimers, N. (2019). Sentence-BERT: Sentence Embeddings using Siamese BERT-Networks. ArXiv Preprint ArXiv:1908.10084.

\bibitem{ribeiro2016} Ribeiro, M. T., Singh, S., \& Guestrin, C. (2016). ``Why should i trust you?'' Explaining the predictions of any classifier. Proceedings of the ACM SIGKDD International Conference on Knowledge Discovery and Data Mining, 13-17-August-2016, 1135--1144.

\bibitem{rosenthal1994} Rosenthal, R., Cooper, H., \& Hedges, L. (1994). Parametric measures of effect size. The Handbook of Research Synthesis, 621(2), 231--244.

\bibitem{russo2018} Russo, D. J., Van Roy, B., Kazerouni, A., Osband, I., \& Wen, Z. (2018). A Tutorial on Thompson Sampling. Foundations and Trends® in Machine Learning, 11(1), 1--96.

\bibitem{sennrich2015} Sennrich, R. (2015). Neural machine translation of rare words with subword units. ArXiv Preprint ArXiv:1508.07909.

\bibitem{singh2024} Singh, C., Inala, J. P., Galley, M., Caruana, R., \& Gao, J. (2024). Rethinking Interpretability in the Era of Large Language Models. https://arxiv.org/abs/2402.01761v1

\bibitem{spivey2006} Spivey, M. J., \& Dale, R. (2006). Continuous dynamics in real-time cognition. Current Directions in Psychological Science, 15(5), 207--211.

\bibitem{strogatz2018} Strogatz, S. H. (2018). Nonlinear dynamics and chaos: with applications to physics, biology, chemistry, and engineering. CRC press.

\bibitem{sussillo2013} Sussillo, D., \& Barak, O. (2013). Opening the black box: low-dimensional dynamics in high-dimensional recurrent neural networks. Neural Computation, 25(3), 626--649.

\bibitem{tegmark2008} Tegmark, M. (2008). The mathematical universe. Foundations of Physics, 38(2), 101--150.

\bibitem{wang2022} Wang, X., Liu, K., Wang, D., Wu, L., Fu, Y., \& Xie, X. (2022). Multi-level Recommendation Reasoning over Knowledge Graphs with Reinforcement Learning. WWW 2022 - Proceedings of the ACM Web Conference 2022, 2098--2108.

\bibitem{weinberg1995} Weinberg, S. (1995). The quantum theory of fields (Vol. 2). Cambridge university press.

\bibitem{welbl2018} Welbl, J., Stenetorp, P., \& Riedel, S. (2018). Constructing Datasets for Multi-hop Reading Comprehension Across Documents. Transactions of the Association for Computational Linguistics, 6, 287--302.

\bibitem{wigner1990} Wigner, E. P. (1990). The Unreasonable Effectiveness of Mathematics in the Natural Sciences. Mathematics and Science, 291--306.

\bibitem{worrall1989} Worrall, J. (1989). Structural Realism: The Best of Both Worlds? Dialectica, 43(1--2), 99--124.

\bibitem{wu2020} Wu, T., Fischer, I., Chuang, I. L., \& Tegmark, M. (2020). Learnability for the information bottleneck. Uncertainty in Artificial Intelligence, 1050--1060.

\bibitem{yang2018} Yang, Z., Qi, P., Zhang, S., Bengio, Y., Cohen, W. W., Salakhutdinov, R., \& Manning, C. D. (2018). HOTPOTQA: A Dataset for Diverse, Explainable Multi-hop Question Answering. ArXiv Preprint.

\bibitem{young1998} Young, N. (1988). An introduction to Hilbert space. Cambridge university press.

\bibitem{zabelina2016} Zabelina, D. L., \& Andrews-Hanna, J. R. (2016). Dynamic network interactions supporting internally-oriented cognition. Current Opinion in Neurobiology, 40, 86--93.

\bibitem{zhang2020} Zhang, W. E., Sheng, Q. Z., Alhazmi, A., \& Li, C. (2020). Adversarial Attacks on Deep-learning Models in Natural Language Processing. ACM Transactions on Intelligent Systems and Technology (TIST), 11(3).

\end{thebibliography}
\end{document}